\documentclass[]{fairmeta}

\usepackage[table,dvipsnames]{xcolor}

\usepackage{microtype}
\usepackage{csquotes}
\usepackage{afterpage}
\usepackage{xspace}
\usepackage{enumitem}
\usepackage{nicefrac}

\usepackage{amsmath}
\usepackage{amssymb}
\usepackage{mathtools}
\usepackage{amsthm}

\usepackage{adjustbox}
\usepackage{booktabs}
\usepackage{colortbl}
\usepackage{hhline}
\usepackage{multirow}
\usepackage{threeparttable}
\usepackage{graphicx}
\usepackage{wrapfig}

\usepackage{etoolbox}
\usepackage{caption}

\definecolor{GroupHeaderGray}{gray}{0.95}
\definecolor{RowLightBlue}{HTML}{E8F4FF}
\definecolor{baselinecolor}{gray}{.9}
\definecolor{tabbeige}{HTML}{F3EAD1}
\definecolor{tabblue}{HTML}{DDE7FF}
\definecolor{tabgreen}{HTML}{DCE9A6}
\definecolor{fbApp}{HTML}{c8e7fa}
\definecolor{fbPurple3}{HTML}{f0ebf5}

\definecolor{deemph}{gray}{0.6}
\definecolor{citecolor}{HTML}{0071BC}
\definecolor{linkcolor}{HTML}{ED1C24}

\definecolor{darkbg}{RGB}{30,30,30}

\newbool{darkmode}
\boolfalse{darkmode}

\newcommand{\adaptivetext}[1]{%
    \ifbool{darkmode}{\textcolor{black}{#1}}{#1}%
}

\newcommand{\adaptivebold}[1]{%
    \ifbool{darkmode}{\textcolor{black}{\textbf{#1}}}{\textbf{#1}}%
}

\newcolumntype{x}[1]{>{\centering\arraybackslash}p{#1pt}}
\newcolumntype{y}[1]{>{\raggedright\arraybackslash}p{#1pt}}
\newcolumntype{z}[1]{>{\raggedleft\arraybackslash}p{#1pt}}

\newlength\savewidth

\renewcommand{\paragraph}[1]{\vspace{1.25mm}\noindent\textbf{#1}}

\newcommand{\app}{\raise.17ex\hbox{$\scriptstyle\sim$}}

\makeatletter
\newcommand{\ssymbol}[1]{$^{\@fnsymbol{#1}}$}
\makeatother

\newcommand{\ours}{OV-Encoder\xspace}
\newcommand{\ourscodec}{OV-Encoder (Codec)\xspace}
\newcommand{\oursframe}{OV-Encoder (Frame)\xspace}

\definecolor{main}{HTML}{2A6175}
\definecolor{Secondary}{HTML}{7FB0A3}
\newcommand{\mainnum}[1]{\textcolor{main}{#1}}
\newcommand{\secondarynum}[1]{\textcolor{Secondary}{#1}}
\newcommand{\baselinenum}[1]{\textcolor{black!55}{#1}}
\definecolor{MolmoPink}{HTML}{F7A0C9}

\newcommand{\multimodalbenchmarktable}{
\begin{table*}[t]
\centering
\setlength{\tabcolsep}{6pt}
\renewcommand{\arraystretch}{1.1}

\caption{\textbf{Comparison of different vision encoders on multimodal benchmarks.}
All models are evaluated on a unified multimodal setting using Qwen3-4B-Instruct2507 as the language backbone.
OV-Encoder-Lang denotes the language-aligned variant of the OV-Encoder architecture.
Qwen3-ViT is extracted from Qwen3-VL-4B.
SigLIP2 uses \texttt{siglip2-so400m-patch16-naflex}. \mainnum{\textbf{(Codec)}} indicates codec-guided visual encoding using motion vectors and residual signals,
while \baselinenum{\textbf{(Frame)}} indicates frame-based visual encoding with dense spatial patchification.
Bold values indicate the best performance under the same evaluation setting.
Results reported in the left columns correspond to encoders trained with caption supervision,
whereas results in the right columns correspond to encoders trained without caption supervision.}
\label{tab:multimodal_benchmarks}

\adjustbox{max width=\textwidth}{
\begin{tabular}{llcc|ccc}
\toprule
\multirow{2}{*}{\textbf{Task}} & \multirow{2}{*}{\textbf{Benchmark}} &
\multicolumn{5}{c}{\textbf{Qwen3-4B-Instruct2507}} \\
\cmidrule(lr){3-7}
& & \textbf{OV-Encoder-Lang} & \textbf{Qwen3-ViT} & \textbf{OV-Encoder} & \textbf{OV-Encoder-Frame} & \textbf{SigLIP2} \\
& & \mainnum{\textbf{(Codec)}} & \baselinenum{\textbf{(Frame)}} & \mainnum{\textbf{(Codec)}} & \baselinenum{\textbf{(Frame)}} & \baselinenum{\textbf{(Frame)}} \\
\midrule

\multirow{7}{*}{\textbf{Video}}
& MVBench
& \mainnum{\textbf{53.2}}
& \baselinenum{47.4}
& \mainnum{\textbf{52.4}}
& \secondarynum{49.8}
& \baselinenum{47.2} \\

& MLVU-dev
& \mainnum{\textbf{47.4}}
& \baselinenum{47.2}
& \mainnum{46.3}
& \secondarynum{\textbf{49.4}}
& \baselinenum{48.4} \\

& NExT-QA (MC)
& \mainnum{\textbf{76.1}}
& \baselinenum{70.1}
& \mainnum{\textbf{75.6}}
& \secondarynum{71.9}
& \baselinenum{70.6} \\

& VideoMME
& \mainnum{\textbf{54.1}}
& \baselinenum{47.2}
& \mainnum{\textbf{53.4}}
& \secondarynum{49.3}
& \baselinenum{46.8} \\

& Perception Test
& \mainnum{\textbf{60.6}}
& \baselinenum{57.1}
& \mainnum{\textbf{60.3}}
& \secondarynum{56.7}
& \baselinenum{56.0} \\

& TOMATO
& \mainnum{21.8}
& \baselinenum{\textbf{22.2}}
& \mainnum{22.2}
& \secondarynum{21.8}
& \baselinenum{\textbf{22.3}} \\

& LongVideoBench-Val-Video
& \mainnum{\textbf{51.6}}
& \baselinenum{45.0}
& \mainnum{\textbf{50.4}}
& \secondarynum{45.5}
& \baselinenum{45.2} \\

\midrule
\multirow{9}{*}{\textbf{Image}}

& AI2D
& \mainnum{\textbf{80.2}}
& \baselinenum{77.8}
& \mainnum{75.7}
& \secondarynum{76.5}
& \baselinenum{\textbf{78.6}} \\

& ChartQA
& \mainnum{\textbf{80.1}}
& \baselinenum{79.6}
& \mainnum{76.5}
& \secondarynum{\textbf{77.8}}
& \baselinenum{76.4} \\

& DocVQA
& \mainnum{83.2}
& \baselinenum{\textbf{85.1}}
& \mainnum{78.4}
& \secondarynum{\textbf{79.5}}
& \baselinenum{75.0} \\

& InfoVQA
& \mainnum{\textbf{51.6}}
& \baselinenum{49.0}
& \mainnum{43.1}
& \secondarynum{\textbf{45.5}}
& \baselinenum{42.0} \\

& MMBench-EN
& \mainnum{\textbf{80.2}}
& \baselinenum{79.4}
& \mainnum{77.2}
& \secondarynum{78.5}
& \baselinenum{\textbf{79.6}} \\

& OCRBench
& \mainnum{657}
& \baselinenum{\textbf{706}}
& \mainnum{605}
& \secondarynum{\textbf{630}}
& \baselinenum{621} \\

& OCRBench v2
& \mainnum{\textbf{30.8}}
& \baselinenum{30.6}
& \mainnum{\textbf{26.3}}
& \secondarynum{26.1}
& \baselinenum{26.1} \\

& MMStar
& \mainnum{\textbf{56.6}}
& \baselinenum{56.6}
& \mainnum{52.1}
& \secondarynum{54.3}
& \baselinenum{\textbf{55.0}} \\

& RealWorldQA
& \mainnum{\textbf{66.1}}
& \baselinenum{63.3}
& \mainnum{60.8}
& \secondarynum{61.2}
& \baselinenum{\textbf{62.1}} \\

\bottomrule
\end{tabular}
}
\end{table*}
}

\newcommand{\videobenchmarktable}{
\begin{table*}[!t]
    \centering
    \setlength{\tabcolsep}{8.5pt}
    \renewcommand{\arraystretch}{1}
    
    \caption{
    \textbf{Comparison with state-of-the-art methods on video understanding benchmarks.}
    We report top-1 accuracy (\%) using an attentive probe with frozen backbones, evaluated under two input configurations:
    \textbf{8 Frames / 2048 Patches} and \textbf{16 Frames / 4096 Patches}.
    For \ourscodec, inputs are constructed using \emph{Dense Video-codec Patchification}, which selectively encodes temporally salient patches from dense video inputs under the corresponding patch budgets.
    \textbf{Bold} indicates the best performance and \underline{underline} indicates the second-best.
    }
    \label{tab:video_benchmarks}
    
    \adjustbox{max width=\textwidth}{
    \begin{tabular}{llc @{\hspace{6mm}} c *{8}{c}}
        \toprule
        \multicolumn{3}{c}{\textbf{Model Setup}} &
        \multicolumn{9}{c}{\textbf{Video Benchmarks (Acc. \%)}} \\
        \cmidrule(lr){1-3} \cmidrule(lr){4-12}
        \textbf{Method} & \textbf{Backbone} & \textbf{Res.} &
        \textbf{Avg.} &
        \rotatebox{90}{SSV2} &
        \rotatebox{90}{Diving48} &
        \rotatebox{90}{Perception Test} &
        \rotatebox{90}{CharadesEgo} &
        \rotatebox{90}{Epic-Verb} &
        \rotatebox{90}{Epic-Noun} &
        \rotatebox{90}{Kinetics-400} &
        \rotatebox{90}{HMDB51} \\
        \midrule
        
        \multicolumn{12}{l}{\textit{8 Frames / 2048 Patches}} \\
        CLIP & ViT-L/14 & 224
        & \baselinenum{50.5} & \baselinenum{48.2} & \baselinenum{46.6} & \baselinenum{52.2} & \baselinenum{10.8} & \baselinenum{52.8} & \baselinenum{36.1} & \baselinenum{79.3} & \baselinenum{78.0} \\
        
        SigLIP & ViT-L/16 & 256
        & \baselinenum{50.1} & \baselinenum{50.7} & \baselinenum{43.9} & \baselinenum{48.9} & \baselinenum{10.9} & \baselinenum{52.2} & \baselinenum{39.1} & \baselinenum{78.2} & \baselinenum{77.0} \\
        
        MetaCLIP & ViT-L/14 & 224
        & \baselinenum{48.5} & \baselinenum{50.6} & \baselinenum{28.9} & \baselinenum{49.8} & \baselinenum{10.4} & \baselinenum{54.1} & \baselinenum{37.1} & \baselinenum{79.6} & \baselinenum{77.1} \\
        
        MetaCLIP2 & ViT-L/14 & 224
        & \baselinenum{50.2} & \baselinenum{47.2} & \baselinenum{48.0} & \baselinenum{47.7} & \baselinenum{11.0} & \baselinenum{48.0} & \baselinenum{40.9} & \baselinenum{82.4} & \baselinenum{76.3} \\
        
        AIMv2 & ViT-L/14 & 224
        & \baselinenum{53.8} & \baselinenum{55.1} & \baselinenum{43.6} & \baselinenum{55.1} & \baselinenum{12.0} & \baselinenum{56.6} & \baselinenum{45.6} & \baselinenum{81.1} & \baselinenum{81.3} \\
        
        SigLIP2 & ViT-L/16 & 256
        & \baselinenum{53.1} & \baselinenum{52.6} & \baselinenum{50.1} & \baselinenum{52.7} & \baselinenum{11.6} & \baselinenum{54.2} & \baselinenum{43.8} & \baselinenum{80.9} & \baselinenum{79.1} \\
        
        DINOv3 & ViT-L/14 & 224
        & \baselinenum{58.0} & \baselinenum{57.4} & \baselinenum{\underline{58.6}} & \baselinenum{\underline{59.3}} & \baselinenum{\adaptivebold{13.2}} & \baselinenum{\adaptivebold{62.5}} & \baselinenum{51.7} & \baselinenum{82.9} & \baselinenum{78.6} \\
        
        \mainnum{\textbf{\oursframe}} & ViT-L/14 & 224
        & \mainnum{\underline{58.4}} & \mainnum{\underline{57.7}} & \mainnum{57.6} & \mainnum{58.3} & \mainnum{12.1} & \mainnum{61.4} & \mainnum{\underline{52.5}} & \mainnum{\underline{84.3}} & \mainnum{\underline{83.1}} \\
        
        \mainnum{\textbf{\ourscodec}} & ViT-L/14 & 224
        & \mainnum{\adaptivebold{60.2}} & \mainnum{\adaptivebold{58.5}} & \mainnum{\adaptivebold{67.2}} & \mainnum{\adaptivebold{60.0}} & \mainnum{\underline{12.3}} & \mainnum{\underline{62.3}} & \mainnum{\adaptivebold{53.9}} & \mainnum{\adaptivebold{84.4}} & \mainnum{\adaptivebold{83.4}} \\
        
        \midrule
        \multicolumn{12}{l}{\textit{16 Frames / 4096 Patches}} \\
        
        SigLIP & ViT-L/16 & 256
        & \baselinenum{52.8} & \baselinenum{52.7} & \baselinenum{54.7} & \baselinenum{51.0} & \baselinenum{11.7} & \baselinenum{54.1} & \baselinenum{40.2} & \baselinenum{79.1} & \baselinenum{78.8} \\
        
        MetaCLIP2 & ViT-L/14 & 224
        & \baselinenum{51.0} & \baselinenum{49.3} & \baselinenum{42.1} & \baselinenum{51.1} & \baselinenum{11.2} & \baselinenum{49.2} & \baselinenum{43.2} & \baselinenum{84.0} & \baselinenum{78.2} \\
        
        AIMv2 & ViT-L/14 & 224
        & \baselinenum{56.4} & \baselinenum{57.2} & \baselinenum{55.7} & \baselinenum{56.4} & \baselinenum{12.4} & \baselinenum{58.3} & \baselinenum{46.2} & \baselinenum{82.2} & \baselinenum{\underline{82.6}} \\
        
        SigLIP2 & ViT-L/16 & 256
        & \baselinenum{55.7} & \baselinenum{58.2} & \baselinenum{56.7} & \baselinenum{53.3} & \baselinenum{11.9} & \baselinenum{56.4} & \baselinenum{45.2} & \baselinenum{82.7} & \baselinenum{81.2} \\
        
        DINOv3 & ViT-L/14 & 224
        & \baselinenum{59.1} & \baselinenum{58.3} & \baselinenum{61.3} & \baselinenum{\underline{60.8}} & \baselinenum{\adaptivebold{14.0}} & \baselinenum{\underline{63.2}} & \baselinenum{51.9} & \baselinenum{83.9} & \baselinenum{79.7} \\
        
        \mainnum{\textbf{\oursframe}} & ViT-L/14 & 224
        & \mainnum{\underline{59.9}} & \mainnum{\underline{58.7}} & \mainnum{\underline{63.2}} & \mainnum{60.3} & \mainnum{12.6} & \mainnum{62.9} & \mainnum{\adaptivebold{54.5}} & \mainnum{\underline{85.1}} & \mainnum{81.6} \\
        
         \mainnum{\textbf{\ourscodec}} & ViT-L/14 & 224
        & \mainnum{\adaptivebold{61.5}} & \mainnum{\adaptivebold{60.1}} & \mainnum{\adaptivebold{69.4}} & \mainnum{\adaptivebold{60.9}} & \mainnum{\underline{12.9}} & \mainnum{\adaptivebold{63.3}} & \mainnum{\underline{54.4}} & \mainnum{\adaptivebold{85.4}} & \mainnum{\adaptivebold{85.3}} \\
        
        \bottomrule
    \end{tabular}
    }
\end{table*}
}

\newcommand{\ovencoderstagetable}{
    \begin{table*}[t]
        \centering
        \setlength{\tabcolsep}{6pt}
        \renewcommand{\arraystretch}{1.1}
        
        \caption{\textbf{Comparison of OV-Encoder training stages on image understanding benchmarks.}
        In this study, we evaluate two training variants of the same ViT architecture under an identical data scale:
        OV-Encoder-stage1 is trained with image-only data, while OV-Encoder-stage2 continues training from Stage~1
        and incorporates OCR and video data with codec-style patch selection.
        \textbf{Bold} values indicate the best performance among the compared encoders.}
        \label{tab:ov_encoder_stage}
        
        \adjustbox{max width=\textwidth}{
        \begin{tabular}{lccc}
            \toprule
            \textbf{Benchmark} 
            & \textbf{OV-Encoder-stage1} 
            & \textbf{OV-Encoder-stage2} 
            & \textbf{SigLIP2-sig} \\
            \midrule
            AI2D            & \mainnum{73.6}  & \mainnum{74.5}  & \baselinenum{\textbf{77.5}} \\
            ChartQA         & \mainnum{73.6}  & \mainnum{76.2}  & \baselinenum{\textbf{77.0}} \\
            DocVQA          & \mainnum{74.3}  & \mainnum{\textbf{78.5}}  & \baselinenum{74.4} \\
            InfoVQA         & \mainnum{34.7}  & \mainnum{\textbf{41.4}}  & \baselinenum{37.7} \\
            MMBench-EN      & \mainnum{74.7}  & \mainnum{76.3}  & \baselinenum{\textbf{78.1}} \\
            MMStar          & \mainnum{49.3}  & \mainnum{49.7}  & \baselinenum{\textbf{52.6}} \\
            RealWorldQA     & \mainnum{61.8}  & \mainnum{61.3}  & \baselinenum{\textbf{62.2}} \\
            OCRBench        & \mainnum{551.0} & \mainnum{\textbf{601.0}} & \baselinenum{590.0} \\
            \bottomrule
        \end{tabular}
        }
    \end{table*}
}

\newcommand{\onevisionpretrainingtable}{
\begin{table}[t!]
\centering
\setlength{\tabcolsep}{6pt}
\renewcommand{\arraystretch}{1.1}

\ifbool{darkmode}{
    \color{white}
    \arrayrulecolor{white}
    \captionsetup{font={color=white}}
}{}
    \caption{
      \textbf{OneVision-Encoder Pretraining Dataset.}
      The pretraining corpus combines large-scale image and video datasets for unified visual representation learning. 
      Image datasets primarily provide broad visual coverage, while video datasets support temporal modeling and video--language alignment. We use “ExoVideo” to denote large-scale third-person web videos, and “ActionVideo” to denote curated action recognition datasets.}
    \adjustbox{width=\textwidth}{
    \begin{tabular}{lccccc}
    \toprule
    Source & Samples & Type & Modality & Temporal & Curation \\
    \midrule
    LAION-400M~\citep{schuhmann2021laion} & 250M & WebImages & Image & -- & Yes \\
    COYO-700M~\citep{kakaobrain2022coyo-700m} & 400M & WebImages & Image & -- & Yes \\
    OBELICS~\citep{laurenccon2023obelics} & 15M & Documents & Image & -- & Yes \\
    Zero250M~\citep{xie2023ccmb} & 15M & CuratedImages & Image & -- & Yes \\
    ImageNet-21K~\citep{deng2009imagenet} & 14M & Images & Image & -- & Yes \\
    \midrule
    HowTo100M~\citep{miech19howto100m} & 50M & ExoVideo & Video & Short & No \\
    Panda-70M~\citep{chen2024panda70m} & 50M & ExoVideo & Video & Long & Yes \\
    Kinetics-710~\citep{li2022uniformerv2} & 658K & ActionVideo & Video & Short & Yes \\
    SSV2~\citep{goyal2017something} & 221K & ActionVideo & Video & Short & Yes \\
    \bottomrule
    \end{tabular}
    }
    \label{table:onevision_pretraining_dataset}
\end{table}
}

\newcommand{\patchscalingtable}{
\begin{table}[t]
\centering
\setlength{\tabcolsep}{6pt}
\renewcommand{\arraystretch}{1.15}

\caption{
\textbf{Effect of patch budget scaling under attentive probing.}
Patch budgets of 512/1024/2048/4096 correspond to 2/4/8/16 video frames, respectively.
Dense SigLIP2 processes all spatial patches per frame, while \ourscodec selectively retains motion-relevant patches guided by codec-derived temporal signals.
}
\label{tab:patch_scaling}

\adjustbox{max width=\textwidth}{
\begin{tabular}{l c cccccccc}
\toprule
\multicolumn{2}{c}{\textbf{Setting}} & \multicolumn{8}{c}{\textbf{Video Benchmarks (Acc. \%)}} \\
\cmidrule(lr){1-1} \cmidrule(lr){2-10}
& \textbf{AVG} &
\rotatebox{90}{SSV2} &
\rotatebox{90}{Diving48} &
\rotatebox{90}{Perception Test} &
\rotatebox{90}{CharadesEgo} &
\rotatebox{90}{Epic-Verb} &
\rotatebox{90}{Epic-Noun} &
\rotatebox{90}{Kinetics-400} &
\rotatebox{90}{HMDB51} \\
\midrule

\multicolumn{10}{l}{\textbf{Patch Budget = 512}} \\
SigLIP2
& \baselinenum{43.4} & \baselinenum{42.8} & \baselinenum{28.1} & \baselinenum{38.7}
& \baselinenum{10.1} & \baselinenum{42.9} & \baselinenum{37.4}
& \baselinenum{74.5} & \baselinenum{72.4} \\
\mainnum{\textbf{\ourscodec}}
& \mainnum{\textbf{50.1}} & \mainnum{50.0} & \mainnum{46.5} & \mainnum{50.5}
& \mainnum{10.6} & \mainnum{50.3} & \mainnum{41.7}
& \mainnum{76.7} & \mainnum{74.7} \\
\midrule

\multicolumn{10}{l}{\textbf{Patch Budget = 1024}} \\
SigLIP2
& \baselinenum{50.8} & \baselinenum{49.0} & \baselinenum{48.7} & \baselinenum{50.1}
& \baselinenum{10.8} & \baselinenum{50.6} & \baselinenum{42.0}
& \baselinenum{78.8} & \baselinenum{76.6} \\
\mainnum{\textbf{\ourscodec}}
& \mainnum{\textbf{56.2}} & \mainnum{56.6} & \mainnum{54.9} & \mainnum{58.6}
& \mainnum{11.1} & \mainnum{58.2} & \mainnum{48.4}
& \mainnum{81.8} & \mainnum{80.3} \\
\midrule

\multicolumn{10}{l}{\textbf{Patch Budget = 2048}} \\
SigLIP2
& \baselinenum{53.1} & \baselinenum{52.6} & \baselinenum{50.1} & \baselinenum{52.7}
& \baselinenum{11.6} & \baselinenum{54.2} & \baselinenum{43.8}
& \baselinenum{80.9} & \baselinenum{79.1} \\
\mainnum{\textbf{\ourscodec}}
& \mainnum{\textbf{60.2}} & \mainnum{58.5} & \mainnum{67.2} & \mainnum{60.0}
& \mainnum{12.3} & \mainnum{62.3} & \mainnum{53.9}
& \mainnum{84.4} & \mainnum{83.4} \\
\midrule

\multicolumn{10}{l}{\textbf{Patch Budget = 4096}} \\
SigLIP2
& \baselinenum{55.7} & \baselinenum{58.2} & \baselinenum{56.7} & \baselinenum{53.3}
& \baselinenum{11.9} & \baselinenum{56.4} & \baselinenum{45.2}
& \baselinenum{82.7} & \baselinenum{81.2} \\
\mainnum{\textbf{\ourscodec}}
& \mainnum{\textbf{61.5}} & \mainnum{60.1} & \mainnum{69.4} & \mainnum{60.9}
& \mainnum{12.9} & \mainnum{63.3} & \mainnum{54.4}
& \mainnum{85.4} & \mainnum{85.3} \\
\bottomrule
\end{tabular}}
\end{table}
}

\newcommand{\codecselectedmotion}{
\begin{table}[t]
\centering
\setlength{\tabcolsep}{6pt}
\renewcommand{\arraystretch}{1.15}

\caption{
\textbf{Controlled interventions on codec-selected motion patches.}
All settings use identical token budgets and visual content unless otherwise specified.
}
\label{tab:causal_interventions}

\adjustbox{max width=\textwidth}{
\begin{tabular}{l c cccccccc}
\toprule
\multicolumn{2}{c}{\textbf{Setting}} & \multicolumn{8}{c}{\textbf{Video Benchmarks (Acc. \%)}} \\ 
\cmidrule(lr){1-1} \cmidrule(lr){2-10}
 & \textbf{AVG} &
\rotatebox{90}{SSV2} & 
\rotatebox{90}{Diving48} & 
\rotatebox{90}{Perception Test} & 
\rotatebox{90}{CharadesEgo} & 
\rotatebox{90}{Epic-Verb} & 
\rotatebox{90}{Epic-Noun} & 
\rotatebox{90}{Kinetics-400} &
\rotatebox{90}{HMDB51} \\
\midrule

\mainnum{\textbf{\ourscodec}}
& \mainnum{\adaptivebold{61.5}}
& \mainnum{\adaptivebold{60.1}} 
& \mainnum{\adaptivebold{69.4}} 
& \mainnum{\adaptivebold{60.9}} 
& \mainnum{\adaptivebold{12.9}} 
& \mainnum{\adaptivebold{63.3}} 
& \mainnum{\adaptivebold{54.4}} 
& \mainnum{\adaptivebold{85.4}} 
& \mainnum{\adaptivebold{85.3}} \\
\midrule

\textit{Non-motion Patch Replacement (50\%)} \\
\footnotesize (Same Video, Same Position)
& \baselinenum{55.4}
& \baselinenum{52.1} & \baselinenum{55.4} & \baselinenum{54.9} & \baselinenum{11.6}
& \baselinenum{56.3} & \baselinenum{50.2} & \baselinenum{83.1} & \baselinenum{79.4} \\
\midrule

\textit{Counterfactual Motion Replacement (50\%) } \\
\footnotesize (Motion Patch from Other Video)
& \baselinenum{54.9}
& \baselinenum{50.6} & \baselinenum{57.2} & \baselinenum{53.8} & \baselinenum{11.3}
& \baselinenum{55.1} & \baselinenum{49.6} & \baselinenum{82.7} & \baselinenum{79.0} \\
\midrule

\textit{Patch--Position Shuffle} \\
\footnotesize (Content Preserved, Positions Shuffled)
& \baselinenum{48.1}
& \baselinenum{41.8} & \baselinenum{46.3} & \baselinenum{45.1} & \baselinenum{8.7}
& \baselinenum{48.2} & \baselinenum{42.5} & \baselinenum{78.4} & \baselinenum{73.6} \\
\bottomrule
\end{tabular}}
\end{table}
}

\title{OneVision-Encoder: Codec‑Aligned Sparsity as a Foundational Principle for Multimodal Intelligence}

\author{Glint Lab}
\author{AIM for Health Lab}
\author{MVP Lab}

\abstract{ \noindent \textbf{Hypothesis.} Artificial general intelligence is, at its core, a compression problem~\citep{sutskever2023observation}. Effective compression demands resonance: deep learning scales best when its architecture aligns with the fundamental structure of the data. These are the fundamental principles. Yet, modern vision architectures have strayed from these truths: visual signals are highly redundant, while discriminative information, the \textit{surprise}, is sparse. Current models process dense pixel grids uniformly, wasting vast compute on static background rather than focusing on the predictive residuals that define motion and meaning. We argue that to solve visual understanding, we must align our architectures with the information-theoretic principles of video, i.e., Codecs.

\vspace{-0.75em}
\noindent \textbf{Method.} OneVision-Encoder encodes video by compressing predictive visual structure into semantic meaning. By adopting Codec Patchification, OneVision-Encoder abandons uniform computation to focus exclusively on the 3.1\%-25\% of regions rich in signal entropy. To unify spatial and temporal reasoning under irregular token layouts, OneVision-Encoder employs a shared 3D RoPE and is trained with a large-scale cluster discrimination objective over more than one million semantic concepts, jointly capturing object permanence and motion dynamics.

\vspace{-0.75em}
\noindent \textbf{Evidence.} The results validate our core hypothesis: efficiency and accuracy are not a trade-off; they are positively correlated. By resolving the dichotomy between dense grids and sparse semantics, \ours redefines the performance frontier. When integrated into large multimodal models, it consistently outperforms strong vision backbones such as Qwen3-ViT and SigLIP2 across 16 image, video, and document understanding benchmarks, despite using substantially fewer visual tokens and pretraining data. Notably, on video understanding tasks, OneVision-Encoder achieves an average improvement of 4.1\% over Qwen3-ViT. Under attentive probing, it achieves state-of-the-art representation quality, with 17.1\% and 8.1\% Top-1 accuracy improvements over SigLIP2 and DINOv3, respectively, on Diving-48 under identical patch budgets. These results demonstrate that codec-aligned, patch-level sparsity is not an optimization trick, but a foundational principle for next-generation visual generalists, positioning OneVision-Encoder as a scalable engine for universal multimodal intelligence.}

\date{\today}
\metadata[Code]{\url{https://github.com/EvolvingLMMs-Lab/OneVision-Encoder}}
\metadata[Data]{\url{https://github.com/EvolvingLMMs-Lab/OneVision-Encoder/blob/main/docs/data_card.md}}
\metadata[Model]{\url{https://huggingface.co/collections/lmms-lab-encoder/onevision-encoder}}

\begin{document}
\maketitle
\section{Introduction}
\label{sec:intro}

\begin{figure*}[t]
\begin{center}
\includegraphics[width=\textwidth]{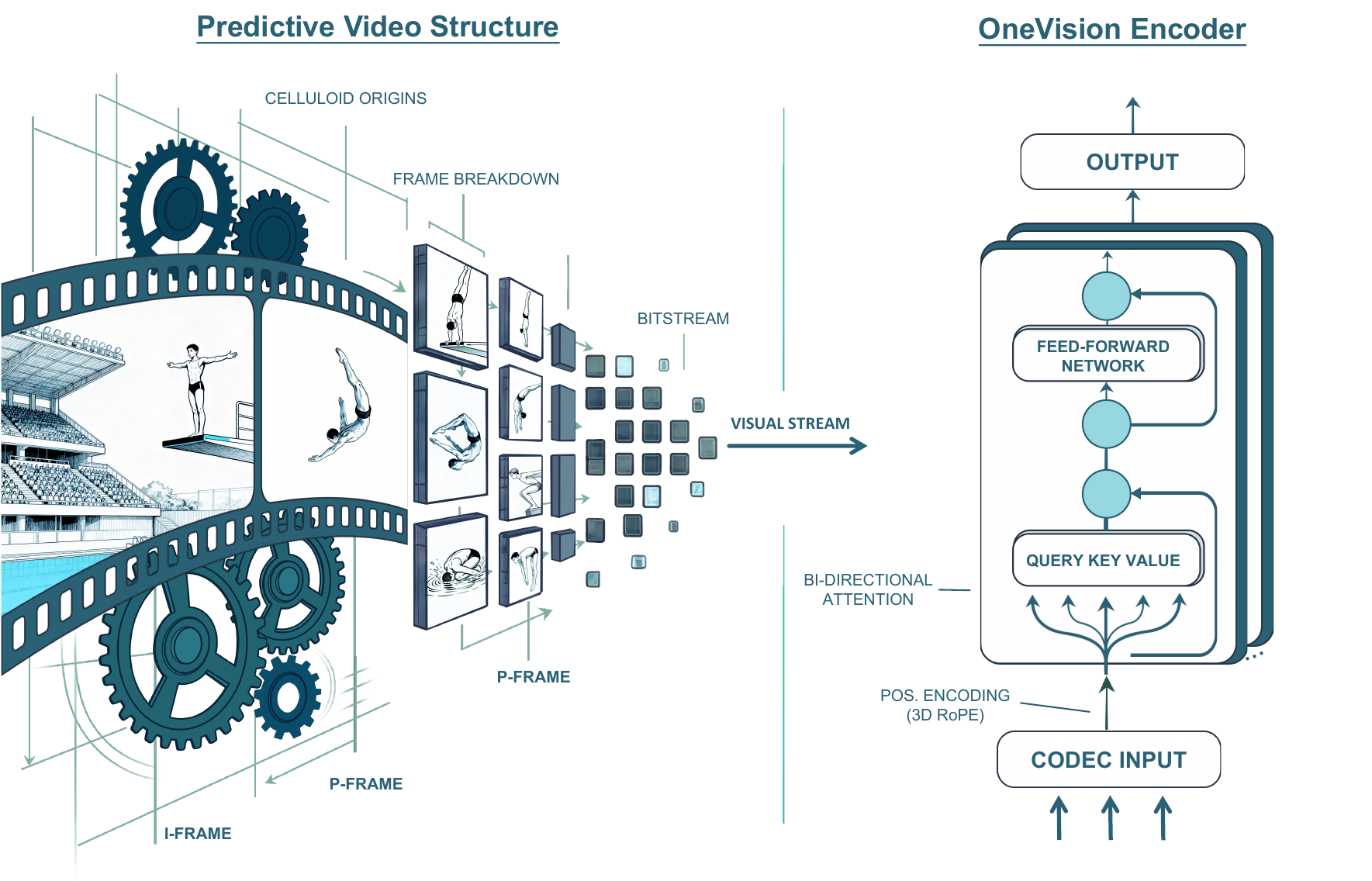}
\caption{\textbf{Visual intelligence as codec-aligned predictive compression.} Visual intelligence as a compression problem, where scalable learning emerges from alignment with the predictive structure of the world. Video exemplifies this principle: most visual content is redundant and predictable, while meaningful information arises sparsely as motion and residual change. Video codecs make this structure explicit by decomposing visual signals into stable spatial context and sparse temporal updates. Grounded in this codec principle, OV-Encoder reframes visual modeling as predictive compression, serving as a scalable engine for universal multimodal intelligence that sees, updates, and reasons over time.}
\label{Patchification}
\end{center}
\end{figure*}

Transformer-based methods have achieved significant improvements in video understanding~\citep{carreira2024scaling,internvideo2,soldan2025residualvit,assran2025v,shu2025video,yang2025streamagent}. By representing videos as sequences of visual tokens, these models have demonstrated a strong capacity to capture long-range spatial and temporal dependencies~\citep{weng2024longvlm,song2025videonsa}. Reconstruction-based self-supervised frameworks (\textit{e.g.,} MAE~\citep{he2021mae}, V-JEPA~\citep{bardes2024revisitingfeaturepredictionlearning}) emphasize pixel- or feature-level prediction, which is effective for capturing low-level spatial and temporal correlations but often lacks explicit semantic structuring. In contrast, contrastive learning paradigms (\textit{e.g.,} CLIP~\citep{clip}, SigLIP~\citep{siglip}) focus on instance-level discrimination and typically rely on external language supervision to induce semantic separation, limiting their ability to model intra-class consistency and fine-grained inter-class relationships. Recent cluster discrimination methods~\citep{an2023unicom,an2024multi,xie2025region,tang2025univit} address this gap by encouraging semantically related entities to form coherent clusters, pointing toward structured object-centric representation learning. Despite these advances, existing video transformers predominantly rely on representations constructed from sparsely sampled frames, leading to dense token sequences that implicitly assume equivalence across spatial regions and temporal frames. Such a frame-centric design reflects a prevailing modeling assumption in video pretraining and motivates rethinking how visual evidence is structured in video signals.

At its core, general intelligence is a compression problem. Natural videos are highly redundant, exhibiting strong spatial and temporal regularities. As a result, the majority of the visual content in a video is predictable from its surrounding context rather than constituting new discriminative evidence. However, standard video pretraining strategies rely on uniform computation over dense pixel grids, expending substantial capacity on static or easily inferred background regions. Discriminative information, the \textit{surprise}, is sparse. This sparsity is not a modeling artifact, but a property of the signal itself. Video compression makes this explicit. Video codecs such as H.264 and H.265/HEVC \textit{(High Efficiency Video Coding)} decompose video signals into spatially complete intra-coded frames (I-frames) that establish global context and predicted frames (P-frames) that encode inter-frame variations via motion compensation and residuals~\citep{sullivan2012overview}. This codec-driven formulation reveals that the vast majority of the video signal corresponds to motion-driven incremental updates to existing spatial context rather than independently discriminative visual evidence. In other words, visual understanding is governed by sparse, localized evidence that defines motion and meaning, rather than dense grids of uniformly processed pixels.

These observations lead to a unified conclusion: to scale visual intelligence, architectures must align with the information-theoretic structure of the data. In this work, we present \textbf{OneVision-Encoder (\ours)}, a HEVC-style Vision Transformer that aligns spatiotemporal representation learning with the intrinsic predictive structure of video signals, as illustrated in Figure~\ref{Patchification}. Rather than uniformly processing dense pixel grids, \ours explicitly determines which visual signals constitute independent evidence and selectively encodes only the regions rich in signal entropy. To enable this, we introduce \textbf{Codec Patchification}, a codec-inspired input formulation that leverages temporal signals exposed by video codecs to organize visual tokens at the patch level, together with a 3D Rotary Position Embedding (RoPE)~\citep{su2024roformer} that jointly encodes spatial and temporal positions to support coherent attention over irregular spatiotemporal layouts. Specifically, \textbf{(a) Dense Video-Codec Patchification}: a codec-inspired video encoding formulation that leverages motion-centric temporal signals exposed by P-frames to patchify selected visual regions (3.1\%-25\%) in dense video inputs, while preserving dense temporal coverage. \textbf{(b) Chunk-wise Patchification}: a codec-inspired temporal patchification scheme that partitions video streams into fixed-length chunks and constructs patch-level representations with chunk-level positional encoding. \textbf{(c) Single-Image Spatial Patchification}: a spatial instantiation of Codec Patchification that constructs patch-level representations for single-image input, enabling structured modeling of static visual content. Furthermore, explicitly modeling visual evidence at the patch level requires a training objective that enforces semantic structure. We adopt a self-supervised cluster discrimination objective based on large-scale semantic clustering over a \textit{concept bank with more than one million clusters}, jointly capturing object-level permanence and motion dynamics. In particular, \ours provides a bi-directional attention-based vision encoder that effectively supports image and video understanding.

Extensive experiments demonstrate the efficacy of \ours across both multimodal and representation-level evaluation protocols. \textbf{For comparison with SigLIP2}~\citep{siglip2}, all models are assessed under identical multimodal fine-tuning conditions, employing a 1.5M-scale LLaVA-Next (single-image instruction)~\citep{liu2024llavanext} and LLaVA-Next-Videos~\citep{zhang2024llavanext-video} instruction-tuning corpus, together with a native-resolution processing strategy. Within this experimental configuration, \ours outperforms SigLIP2 across 16 benchmarks spanning video, image, and document understanding tasks, when evaluated using large multimodal models (LMMs) built upon Qwen3-4B~\citep{Qwen3-VL}.

\textbf{For comparison with Qwen3-ViT}~\citep{Qwen3-VL}, we adopt a controlled evaluation protocol. Specifically, we integrate OneVision-Encoder with the Qwen3-1.7B language model and train it under the LLaVA-OneVision-1.5~\citep{an2025llava} framework, completing both Stage 1 and Stage 1.5 to adapt the encoder to native-resolution inputs. The trained OneVision-Encoder is then decoupled and compared with Qwen3-ViT under the same LLaVA-Next-Videos instruction-tuning training setting, where \ours outperforms Qwen3-ViT across 16 understanding benchmarks under an LMM built upon Qwen3-4B. In particular, despite having been pretrained on substantially fewer visual–text tokens (approximately 100B caption tokens), \ours outperforms Qwen3-ViT, which is pretrained on more than 2.1T caption and instruction-aligned tokens.

Under attentive probing on 7 benchmarks, \ours achieves state-of-the-art performance, including a 17.1\% and 8.1\% Top-1 accuracy improvements over SigLIP2 and DINOv3~\citep{simeoni2025dinov3}, respectively, on Diving-48 under an identical patch budget of 2048. Moreover, \ours outperforms strong vision baselines such as DINOv3, SigLIP2, MetaCLIP2~\citep{chuang2025meta}, and AIMv2~\citep{fini2025multimodal} under dense-patch evaluation. We release our data, training protocols, and model parameters to support transparent, reproducible, and cost-effective vision–language research. Our contributions are as follows:
\begin{itemize}
\item We present OneVision-Encoder (\ours), a HEVC-style vision transformer that aligns spatiotemporal representation learning with the intrinsic predictive structure of video signals through Codec patch-level encoding.

\item We introduce Codec Patchification, a codec-inspired input formulation that leverages codec-derived temporal signals to selectively encode informative visual patches (3.1\%-25\%) from dense video, while unifying video, chunk-wise sampling, and single-image inputs with 3D-RoPE.

\item We adopt a self-supervised cluster discrimination objective that jointly models object-level and motion-level semantics with a large-scale concept bank, enabling structured and modality-agnostic visual representation learning.

\item Extensive experiments establish the effectiveness of \ours across evaluation settings. Under LLM-based probing, the model consistently surpasses strong vision backbones, including Qwen3-ViT and SigLIP2, across multimodal benchmarks. Under attentive probing, \ours outperforms SigLIP2 and DINOv3 across 7 benchmarks, indicating robust representation learning.

\end{itemize}

\section{Approach}
Although most previous video encoders focus on short clips of 16 frames (roughly seconds)~\citep{vjepa,internvideo2}, we explore training with longer clips of up to 64 frames at higher spatial resolutions. Let $\mathcal{V}_i=\{\mathcal{I}_{i,t}\}_{t=1}^{T_i}$ denote the $i$-th input video of spatial size $H\times W$, where $T_i$ is the total number of frames in $\mathcal{V}_i$. Our objective is to process raw video inputs within the proposed different input configurations and jointly encode them using a shared ViT for unified spatiotemporal representation learning.

\subsection{HEVC Guided Patch Selection}

\paragraph{Codec Based Video Factorization.}
Following the standard High Efficiency Video Coding formulation~\citep{sullivan2012overview}, each video $\mathcal{V}_i$ is divided into $N_i$ Groups of Pictures (GOP), $\{\mathcal{S}_{i,n}\}_{n=1}^{N_i}$. Applying the HEVC codec to each Groups of Pictures yields one intra-coded frame and $(K_{i,n}{-}1)$ predicted frames:
\begin{equation}
\label{eq:hevc-op}
\big(F^{\mathrm{I}}_{i,n},\, \{F^{\mathrm{P}}_{i,n,\tau}\}_{\tau=1}^{K_{i,n}-1}\big)
~=~ \mathcal{C}_{\mathrm{HEVC}}(\mathcal{S}_{i,n}),
\end{equation}
where $F^{\mathrm{I}}_{i,n}\in\mathbb{R}^{H\times W\times C_{\text{img}}}$ is the I-frame (RGB, $C_{\text{img}}=3$), and each $F^{\mathrm{P}}_{i,n,\tau}$ denotes a P-frame, which is represented in the bitstream by motion vectors and a residual signal. $K_{i,n}$ denotes the GOP length.

\paragraph{Motion and Residual Signals.}
In HEVC, motion is represented by motion vectors $\boldsymbol{d}_{i,n,\tau}$ that encode block level displacements between the current frame and its motion compensated prediction from the reference frame. Concretely, P-frames are partitioned into coding units (CUs) with variable sizes ranging from $4{\times}4$ to $64{\times}64$, and all pixels within a CU share the same motion vector. To align codec signals with ViT patchification, we first broadcast each CU motion vector to its covered pixels, obtaining a dense pixel level motion field. The magnitude $\|\boldsymbol{d}_{i,n,\tau}\|_2$ reflects the intensity of local motion, with larger values indicating stronger or more complex dynamics. In addition to motion, each P-frame is associated with a residual signal that captures appearance changes not explained by motion compensation; we decode the luma residual into the pixel domain and measure its energy as a complementary cue for unpredictable visual variation. At the patch level, we aggregate motion magnitude and residual energy over all pixels inside each ViT patch, so the two signals jointly characterize the amount of new visual evidence introduced by a region. We use these codec exposed signals as a principled proxy for unpredictability, enabling the identification of salient regions that contribute new spatiotemporal information.

\paragraph{Sparse Patch Selection.}
Let $p$ denote the patch size (\textit{e.g.}, $p{=}14$). We define the patch grid $\mathcal{G}=\{(y,x)\mid 0\!\le\!y\!<\!H/p,\;0\!\le\!x\!<\!W/p\}$, with cardinality $P_0=(H/p)(W/p)$. For each P-frame, we compute a patch level saliency score by aggregating the codec exposed motion magnitude and residual energy defined above. Based on the aggregated saliency score, we construct a binary mask $\Omega_{i,n,\tau}\subseteq\mathcal{G}$ by selecting a fixed proportion of the most salient patches. The sparsity ratio is fixed throughout training and inference by selecting a fixed proportion $r$ of the most salient patches, i.e., $|\Omega_{i,n,\tau}| = \lfloor r P_0 \rfloor$.

\begin{figure*}[!t]
\begin{center}
\includegraphics[width=\textwidth]{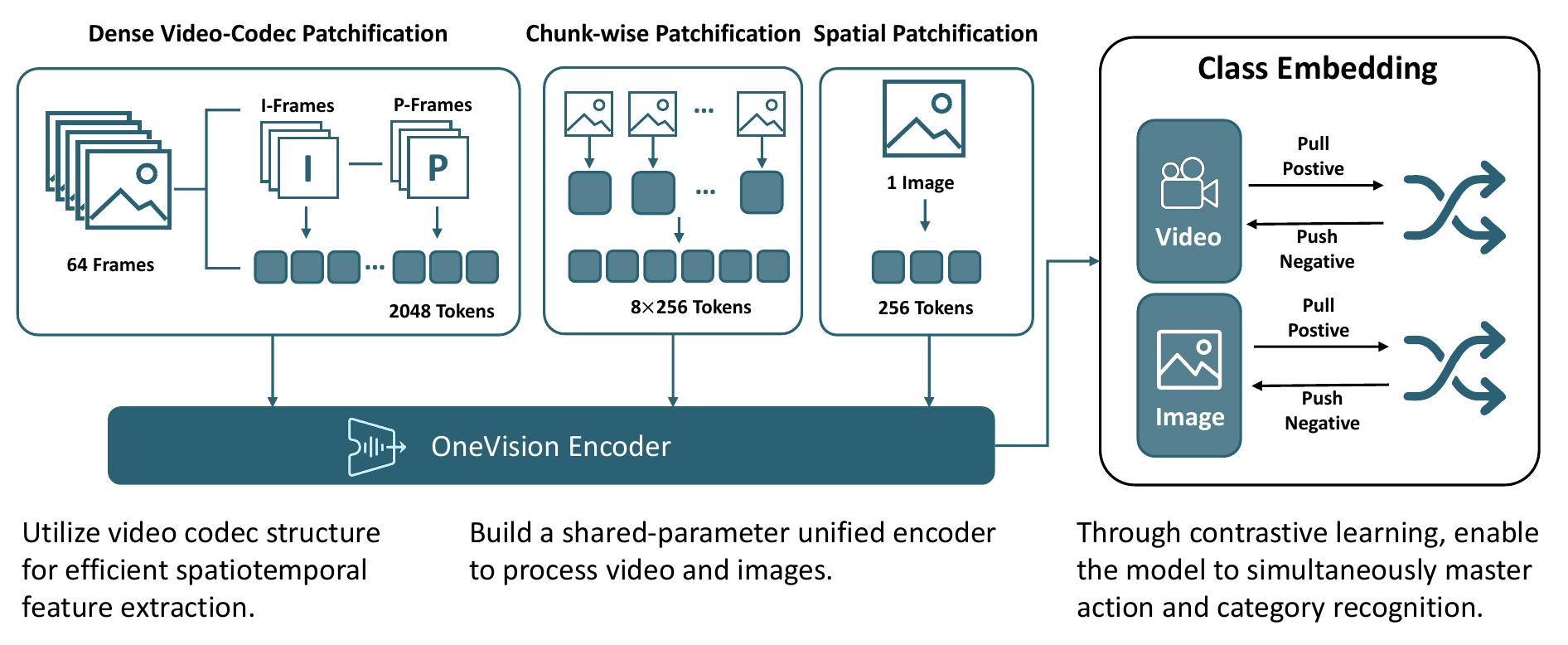}
\caption{\textbf{Overview of the OneVision-Encoder framework.} \textbf{Left:} Input formulation. The framework integrates three Codec Patchification strategies: Dense Video-Codec Patchification, Chunk-wise Patchification, and (Sigle-Image/Frame) Spatial Patchification. All inputs are processed by a shared-parameter OneVision-Encoder. \textbf{Right:} Unified cluster discrimination objective. Image and video embeddings are aligned through contrastive learning against a global set of class centers, jointly optimizing object-level and action-level representations within a single encoder.}
\label{fig:teaser}
\end{center}
\end{figure*}

\subsection{Codec Patchification}

\paragraph{Dense Video-Codec Patchification.}
Following the codec formulation, each video $\mathcal{V}_i$ is partitioned into $N_i$ GOP. For the $n$-th GOP, the HEVC encoder produces one intra-coded frame $F^{\mathrm{I}}_{i,n}$ and $(K_{i,n}-1)$ predicted frames $\{F^{\mathrm{P}}_{i,n,\tau}\}_{\tau=1}^{K_{i,n}-1}$ that are defined by motion vectors and a residual signal after motion compensation. Let $\Omega_{i,n,\tau}$ denote the codec-derived binary mask that selects dynamically informative patches based on motion magnitude and residual energy, and let $\Pi_p(\cdot)$ denote patchification with patch size $p$. The HEVC-compressed input sequence is defined as
\begin{equation}
\label{eq:hevc-input}
\mathcal{F}_{i,n}^{(\mathrm{hevc})}
~\triangleq~
\Big[
  \Pi_p(F^{\mathrm{I}}_{i,n})
  \;\oplus\;
  \big\{\Pi_p(\bar{F}^{\mathrm{P}}_{i,n,\tau})[\Omega_{i,n,\tau}]\big\}_{\tau=1}^{K_{i,n}-1}
\Big],
\end{equation}
where $\bar{F}^{\mathrm{P}}_{i,n,\tau}\in\mathbb{R}^{H\times W\times C_{\text{img}}}$ denotes the decoded RGB P-frame in the pixel domain obtained by decoding the HEVC bitstream.
The binary mask $\Omega_{i,n,\tau}$ is computed from the codec-exposed motion vectors $\boldsymbol{d}_{i,n,\tau}$ and the associated motion-compensated residual signal of the same P-frame, and is used only for salient patch selection.
Therefore, the tokens fed to the encoder for P-frames are RGB patches from $\bar{F}^{\mathrm{P}}_{i,n,\tau}$ indexed by $\Omega_{i,n,\tau}$.
$K_{i,n}$ is the GOP length. $[\Omega]$ denotes masked patch selection along the patch dimension. $\oplus$ denotes concatenation along the token dimension.
Let $M_{i,n}$ be the number of tokens in $\mathcal{F}_{i,n}^{(\mathrm{hevc})}$; the total tokens over the video satisfy:
\begin{equation}
\label{eq:token-count}
M_i = \sum_{n=1}^{N_i} M_{i,n}
~=~ \sum_{n=1}^{N_i}\!\Big(P_0 + \sum_{\tau=1}^{K_{i,n}-1}\!|\Omega_{i,n,\tau}|\Big),
\end{equation}
which yields a pixel/token compression ratio per GOP, $\gamma_{i,n}=1 - M_{i,n}/(K_{i,n}P_0)$.
Under our default setting (64 frames, GOP size 32, token budget 2048, $P_0=256$), the overall clip-level token reduction is $1 - 2048/(64P_0) = 87.5\%$.

\paragraph{Chunk-wise Patchification.}
To unify Codec patchification with sparse temporal sampling, each video $\mathcal{V}_i$ is uniformly partitioned into $C$ temporal chunks of $\lfloor T_i/C \rfloor$ consecutive frames. From every chunk, one frame is randomly sampled, resulting in a temporally stratified subsequence $\pi_i=\{t_c\}_{c=1}^{C}$, where $t_c\!\in\![ (c{-}1)\lfloor T_i/C \rfloor ,\, c\lfloor T_i/C \rfloor )$. The corresponding chunk-wise sampling sequence is defined as
\begin{equation}
\label{eq:flexible-input}
\mathcal{F}_i^{(\mathrm{chunk})}
~\triangleq~
\big\{
  \Pi_p(\mathcal{I}_{i,t_c}) \mid t_c\!\in\!\pi_i
\big\}
~=~
\Pi_p(\mathcal{V}_i[\pi_i]).
\end{equation}
\paragraph{Single-Image Spatial Patchification.}
To achieve spatial scalability, each image $\mathcal{I}_{i,t}\in\mathbb{R}^{H\times W\times C_{\text{img}}}$ is processed independently as a static image input. Each image is directly patchified in a row-wise manner from top to bottom to ensure a deterministic spatial ordering of visual tokens, preserving the original spatial layout. The resulting patch sequence is defined as
\begin{equation}
\mathcal{F}_i^{(\mathrm{image})}
~\triangleq~
\big\{
  \Pi_p(\mathcal{I}_{i,t}) \mid 
  t = 1,\dots,T_i
\big\},
\end{equation}
where $\Pi_p(\cdot)$ denotes the patchification operator applied to a single frame, and $T_i=1$ for single-image inputs. A detailed spatial bias analysis is provided in the supplementary material.

\begin{figure}
    \centering
    \includegraphics[width=\textwidth]{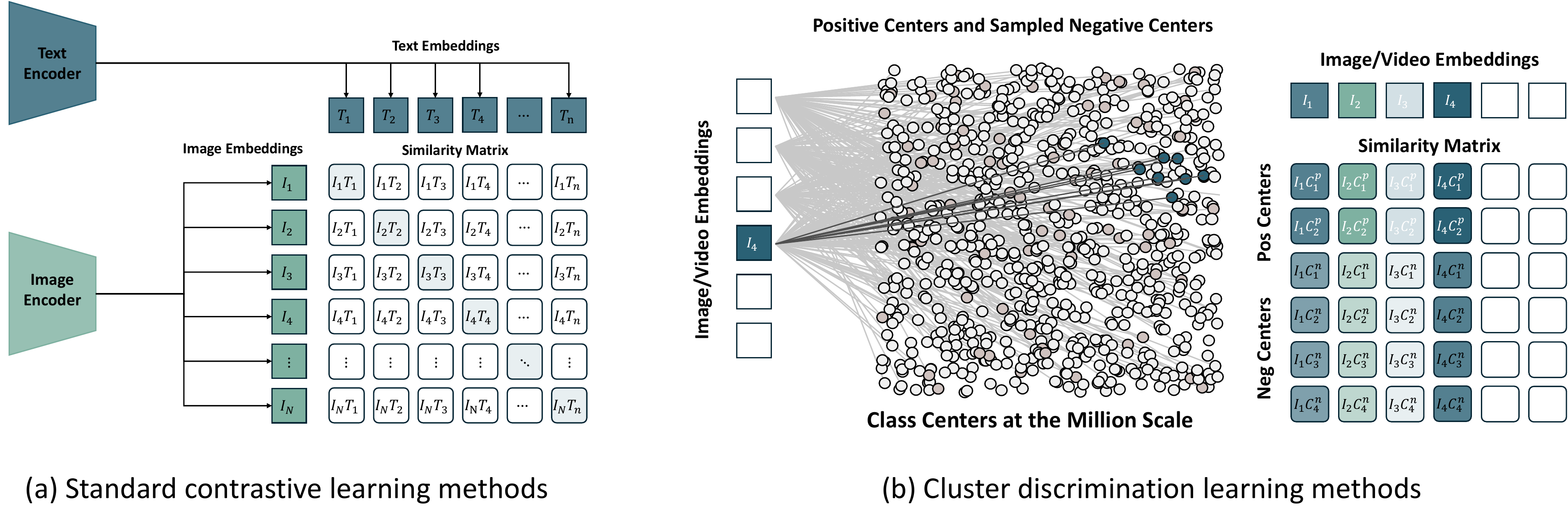}
    \caption{\textbf{Contrastive learning vs. cluster discrimination.} (a) Standard contrastive learning contrasts samples against batch-local negatives, constraining the view of the embedding space. (b) Cluster discrimination contrasts samples against a global concept bank of clustered centers at scale, yielding discriminative and structurally separated representations.}
    \label{fig:mlcd}
\end{figure}

\paragraph{Tokenization and Transformer Encoding.}
After generating the three types of input sequences, namely the Codec Patchification formulation, we uniformly tokenize and encode them using a shared Transformer backbone $\phi(\cdot)$. 
For each video input $\mathrm{vid}\!\in\!\{\mathrm{hevc},\mathrm{chunk},\mathrm{image}\}$, the token sequence $\mathcal{F}_i^{\mathrm{vid}}=\{x_{i,k}^{\mathrm{vid}}\}_{k=1}^{M_i^{\mathrm{vid}}}$ is processed by the encoder $\phi(\cdot)$ followed by a linear projection head $W\!\in\!\mathbb{R}^{d\times D}$ where $d$ and $D$ denote hidden and latent dimensions of the encoder, respectively, to obtain latent embeddings:
\begin{equation}
\label{eq:encoding}
E_i^{\mathrm{vid}}
~=~
\phi \big(\mathcal{F}_i^{\mathrm{vid}}\big)W
~=~
Z_i^{\mathrm{vid}}W.
\end{equation}
which defines $Z_i^{\mathrm{vid}}=\phi(\mathcal{F}_i^{\mathrm{vid}})\in\mathbb{R}^{M_i^{\mathrm{vid}}\times d}$ as the encoder output features before projection. Finally, a function $f(\cdot)$, such as attentive pooling, integrates all token embeddings into a compact video-level representation $e_i^{\mathrm{vid}}=f(E_i^{\mathrm{vid}})\!\in\!\mathbb{R}^{D}$.

\subsection{Image and Video Clustering}

Iterative clustering-discrimination approaches commonly suffer from substantial computational overhead~\citep{caron2018deep}. To address this issue, we adopt a single-step offline clustering to efficiently capture both object-level semantics from images and motion-level semantics from videos. Notably, embeddings used for clustering are extracted using a separate frozen vision encoder (e.g., metaclip-h14~\citep{metaclip})

For the image modality, we follow the clustering formulation in MLCD~\citep{an2024multi} to capture object-level semantics. Given an image embedding $e_i^{\mathrm{obj}} \in \mathbb{R}^D$, we learn a set of object-level semantic centroids $\mathcal{C}_{\mathrm{obj}} = \{c_k^{\mathrm{obj}}\}_{k=1}^{K_{\mathrm{obj}}} \subseteq \mathbb{R}^D$ by minimizing the within-cluster distance:
\begin{equation}
\label{eq:obj_cluster}
\mathcal{C}_{\mathrm{obj}}
~=~
\arg\min_{\{c_k^{\mathrm{obj}}\}}
\sum_{i=1}^{N_{\mathrm{obj}}}
\min_{1\le k\le K_{\mathrm{obj}}}
\big\| e_i^{\mathrm{obj}} - c_k^{\mathrm{obj}} \big\|_2^2 ,
\end{equation}
where $N_{\mathrm{obj}}$ denotes the number of image samples.

For the video modality, we extend this formulation to model motion-level dynamics. Video embeddings are derived from fixed-length 16-frame inputs, where frame-level features are concatenated to form a single video-level representation $e_i^{\mathrm{vid}}\in\mathbb{R}^{D}$. We then define a set of video semantic centroids $\mathcal{C}_{\mathrm{vid}}=\{c_k^{\mathrm{vid}}\}_{k=1}^{K_{\mathrm{vid}}}\subseteq\mathbb{R}^{D}$, and formulate the clustering objective for videos as
\begin{equation}
\label{eq:vid_cluster}
\mathcal{C}_{\mathrm{vid}}
~=~
\arg\min_{\{c_k^{\mathrm{vid}}\}}
\sum_{i=1}^{N_{\mathrm{vid}}}
\min_{1\le k\le K_{\mathrm{vid}}}
\big\| e_i^{\mathrm{vid}} - c_k^{\mathrm{vid}} \big\|_2^2 ,
\end{equation}
where $N_{\mathrm{vid}}$ denotes the number of video samples. We define a set of shared semantic centroids \( \mathcal{C}_{uni} = \{c_k^{\mathrm{obj}}\}_{k=1}^{K_{\mathrm{obj}}} \cup \{c_k^{\mathrm{vid}}\}_{k=1}^{K_{\mathrm{vid}}} \), and \(K = K_{\mathrm{obj}} + K_{\mathrm{vid}}\) represents the total number of clusters across both image and video modalities. These semantic centroids are later used as supervision signals in the cluster discrimination objective, where image embeddings are supervised by object-level centroids and video embeddings are supervised by motion-level centroids.

\begin{figure}
    \centering
    \includegraphics[width=\textwidth]{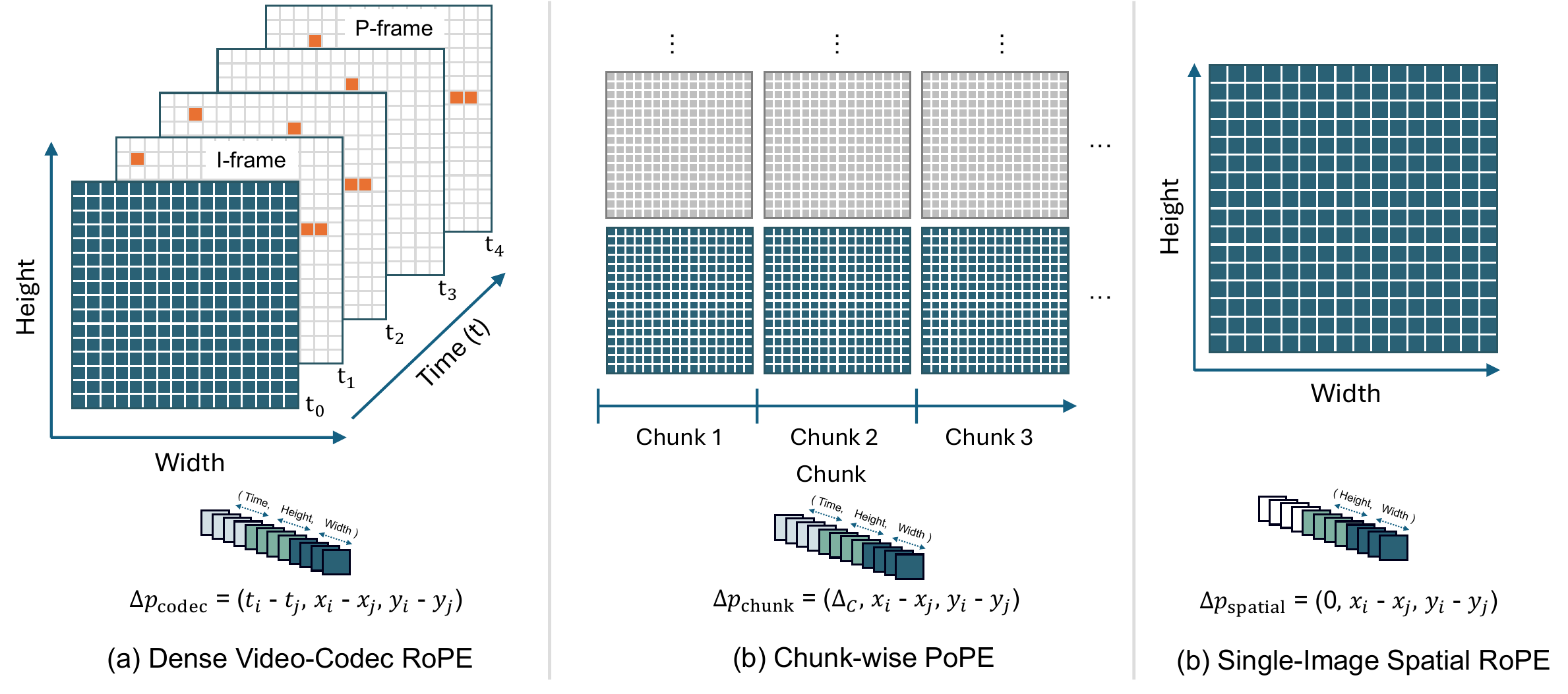}
    \caption{\textbf{3D-RoPE for Codec Patchification.} A unified relative positional encoding scheme is adopted for Codec Patchification. (a) encodes full spatiotemporal offsets $(\Delta t,\Delta x,\Delta y)$ over I/P-frame sequences to preserve motion-driven inter-frame structure. (b) defines temporal offsets at the chunk level, enabling structured reasoning under non-uniform temporal sampling. (c) degenerates the formulation to purely spatial offsets $(0,\Delta x,\Delta y)$ for static inputs. 3D-RoPE preserves structural consistency, enabling coherent attention over sparse and irregular token layouts.}
    \label{fig:rope}
\end{figure}

\subsection{Training Objective}
Visual samples commonly exhibit multiple semantic components, including object-level semantics from images and motion-level semantics from videos, rendering single-label assignments inadequate for unified representation learning. To capture both semantic structures, we introduce a contrastive objective that leverages these semantic clusters as pseudo-label supervision to explicitly enforce structural constraints, as illustrated in Figure~\ref{fig:mlcd}. Specifically, for each visual embedding $e_i \in \mathbb{R}^D$, we identify multiple positive semantic labels from a unified semantic centroid set $\mathcal{C}_{uni}$, which consists of both object-level centroids $\mathcal{C}_{obj}$ and motion-level centroids $\mathcal{C}_{vid}$. We compute the training objective separately for each semantic granularity $m \in \{\mathrm{obj}, \mathrm{vid}\}$, where negative labels are drawn from the corresponding centroid set $\mathcal{C}_m$. The remaining centroids within the same granularity are treated as negative labels. Subsequently, the joint multi-label semantic discrimination objective is formulated as:
\begin{equation}
\mathcal{L}
=\sum_{m\in\{\mathrm{obj},\mathrm{vid}\}}
\mathbb{E}_{(u,k)\sim\mathcal{C}_m}
\log\!\big(1+\exp(-y_{u,k}^m\,\sigma_{u,k}^{m})\big),
\end{equation}
where $\sigma_{u,k}^{m}=e_u^{\top}c_k^{m}$ denotes the similarity score between the embedding $e_u$ and its semantic centroid $c_k^{m}$, and $u$ indexes visual embeddings while $k$ indexes centroids in $\mathcal{C}_m$. For each granularity $m$, $(u,k)\!\sim\!\mathcal{C}_m$ samples visual embeddings $e_u$ and semantic centroids $c_k^{m}$ from the corresponding cluster set, with pseudo-labels $y_{u,k}^m\!\in\!\{+1,-1\}$ indicating positive or negative semantic associations. This unified formulation integrates both object- and motion-level clustering signals, enforcing spatiotemporal consistency and promoting discriminative representation learning.

\subsection{Architecture}
For the OneVision-Encoder, we adopt the Vision Transformer (ViT) architecture~\citep{dosovitskiy2020vit}. 

\noindent\textbf{3D-RoPE for Codec Patchification.}
Unlike absolute encoding \(p=(t,x,y)\), 3D-RoPE adopts a relative positional scheme~\citep{su2024roformer} represented as $\Delta p=(t_1-t_2, x_1-x_2, y_1-y_2)$, as illustrated in Figure~\ref{fig:rope}. The relative offsets \(\Delta p\) for the three inputs are defined as:
\begin{itemize}[leftmargin=12pt]
\item \emph{Dense Video-Codec Patchification}: \(\displaystyle \Delta p_{\text{codec}}=(t_i{-}t_j,\;x_i{-}x_j,\;y_i{-}y_j)\), which emphasizes inter-frame (I/P) residual alignment via the temporal offset \(t_i{-}t_j\).
\item \emph{Chunk-wise Patchification}: for frames $t_i,t_j$ belonging to chunks $c_i,c_j\!\in\!\{1,\dots,C\}$, the relative offset is defined as \(\displaystyle \Delta p_{\text{chunk}} =(\Delta c,\;\Delta x,\;\Delta y), \) capturing inter-chunk temporal disparity under non-uniform sampling, where $\Delta c = c_i - c_j$.
\item \emph{Single-Image Spatial Patchification}: for two patches within the same frame $t$ with spatial coordinates $(x_i, y_i)$ and $(x_j, y_j)$, the relative positional offset is defined as
\(
\Delta p_{\text{spatial}}
=(0,\;\Delta x,\;\Delta y),
\)
where $\Delta x = x_i - x_j$ and $\Delta y = y_i - y_j$, encoding spatial positional relationships without temporal shifts.
\end{itemize}

\noindent\textbf{Attentive Pooling Head.} We employ a multi-head attention pooling module, adapted from SigLIP~\citep{siglip}, to aggregate spatiotemporal tokens into compact class embeddings through learnable token-to-class attention weights, emphasizing salient regions and enabling unified global contextual representation across image and video modalities.

\section{Pretraining Dataset}
\label{subsec:pretraining_dataset}

\paragraph{\bf Data Annotation and Processing.}
This section details the annotation and processing pipeline for the OneVision-Encoder pretraining data. Our core objective is to generate high-quality supervision signals for massive-scale data through automated means.

\noindent\textbf{Image Data Annotation.} For image data, we primarily process LAION-400M~\citep{laion} and COYO-700M~\citep{kakaobrain2022coyo-700m}. First, we employ a Union-Find algorithm to strictly deduplicate the dataset. Subsequently, we utilize the metaclip-h14-fullcc2.5b~\citep{metaclip} model to extract image features and cluster all images into two million classes. Based on this clustering, each image sample is annotated with the nearest Top-10 class centers as its multi-label supervision signal. Furthermore, we incorporate the OBELICS~\citep{laurenccon2023obelics} and Zero250M~\citep{xie2023ccmb} datasets. We utilize PaddleOCR~\citep{sarkar2024automatic} to recognize text within images and perform word segmentation on the recognized content; the resulting vocabulary is used as multi-labels to construct a supervision signal containing exactly 100 fine-grained tags per image.

\noindent\textbf{Video Data Annotation.} The video data construction encompasses HowTo100M~\citep{miech19howto100m}, Panda-70M~\citep{chen2024panda70m}, Kinetics-710~\citep{li2022uniformerv2}, and Something-Something-V2 (SSV2)~\citep{goyal2017something}. We uniformly adopt metaclip-h14-fullcc2.5b as the video encoder, performing uniform frame sampling to extract features from a fixed 8-frame clip. In the feature processing stage, we adopt a ``L2 Normalize  $ \rightarrow $  Concatenate  $ \rightarrow $ L2 Normalize'' strategy to generate video-level representations. We then cluster the video representations into 400k classes and assign each video clip the nearest Top-10 class centers as its multi-labels.

\onevisionpretrainingtable

\section{Experiments}

\subsection{Pretraining OneVision-Encoder}
\label{subsec:pretraining-ove}

We adopt a two-stage pretraining pipeline using large-scale image, video, and OCR data, trained on 128 A800 GPUs (16 nodes × 8 GPUs).

\paragraph{Stage 1:} 
For the image model, we use images at a resolution of 224. We adopt the AdamW optimizer with a learning rate of 0.001 and a weight decay of 0.2. The number of classes (k) is two million, the ratio of sampled negative class centers (r) is 0.1, and the number of positive labels (l) assigned to each image (and each video) is 10. In this stage, we trained on 13B samples using only image data.

\paragraph{Stage 2:} 
In the second pre-training stage, we introduced OCR data and video data, training on 4B samples with an image resolution of 448 and video resolution of 224 (frame sampling). For video processing, we randomly adopted one of Codec Patchification formulations. For dense video-Codec inputs, each training sample corresponds to a fixed-length clip of 64 consecutive frames. We follow an HEVC-style GOP configuration with an I-frame every 32 frames, resulting in two I-frames and sixty-two P-frames per clip. All I-frames are fully encoded, retaining all spatial patches to establish complete spatial context. For the remaining P-frames, codec-derived motion vectors and residuals are used to identify temporally salient regions, and patches are selected sparsely across all P-frames.  Importantly, the patch budget is enforced at the clip level rather than the GOP level. While GOPs define the I/P frame structure, the total number of visual tokens for the entire 64-frame clip is fixed to 2,048. Specifically, 512 tokens are allocated to the two I-frames (256 patches each), and the remaining 1,536 tokens are distributed across all P-frames by selecting motion-relevant residual patches. This results in an 87.5\% reduction compared to dense processing of all 64 frames (16,384 patches), while preserving full temporal coverage. All samples shared the same ViT backbone, with loss calculated separately for each modality based on their annotations. The video-to-image ratio was 1:1, and the learning rate was reduced to 0.0001.

\noindent\textbf{Training Strategy.} During the training phase, all data sources share the same OneVision-Encoder for feature extraction. However, when computing the loss, the loss is calculated separately for each data type and then aggregated. Detailed training configurations and implementation specifics are provided in the supplementary material.

\subsection{LMM Probing Evaluation - Language Alignment}
In this section, we evaluate the effectiveness of OV-Encoder when integrated into LMMs. Our goal is to assess whether the learned visual representations transfer robustly to multimodal reasoning settings, while isolating the contribution of the vision encoder from language model capacity and training protocol. We lift these features through alignment tuning to construct a new codec encoder, \textbf{OV-Encoder-Lang}, specialized for MLLMs. The controlled evaluation pipeline is illustrated in Figure~\ref{fig:LLMProbing} in the supplementary material.

\textbf{MLLM Evaluation Tasks.} We adapt vision encoders to large multimodal models and evaluate downstream performance on a set of image- and video-centric benchmarks using LMMs-Eval~\citep{lmms_eval} with the default prompt. \textbf{Image tasks} include ChartQA~\citep{chartqa}, DocVQA~\citep{docvqa}, InfoVQA~\citep{infovqa}, AI2D~\citep{ai2d}, MMBench-EN~\citep{mme}, OCRBench~\citep{ocrbench}, OCRBench v2~\citep{ocrbench}, MMStar~\citep{mmstar}, and RealWorldQA~\citep{realworldqa}. \textbf{Video tasks} include MVBench~\citep{mvbench}, MLVU-dev~\citep{mlvu}, NExT-QA~\citep{nextqa}, VideoMME~\citep{videomme}, PerceptionTest~\citep{perceptiontest}, TOMATO~\citep{shinoda2025tomato}, and LongVideoBench-Val-Video~\citep{longvideobench}. 

\paragraph{Experimental Setup.}
Following a probing-based fine-tuning paradigm, we keep the language model architecture fixed and vary only the visual encoder. Unless otherwise specified, all experiments are conducted with Qwen3-4B-Instruct2507 as the language backbone. We adopt a two-stage training pipeline: first, a Stage-1 alignment phase, followed by a Stage-2 instruction-tuning phase. The instruction-tuning corpus consists of approximately 1.5M samples, including 740K single-image instruction from LLaVA-Next and 800K samples from LLaVA-Next-Videos. This unified instruction-following setting enables a controlled comparison across different vision encoders under identical multimodal supervision.

\subsubsection{Native-Resolution Evaluation.}
We adopt a native-resolution evaluation strategy following LLaVA-Next, with a key distinction: input frames matching the native resolution of the vision encoder are processed directly without spatial tiling or cropping. For video inputs, we use a fixed per-frame resolution, set to $512 \times 512$ for SigLIP2 and Qwen3-ViT, and $504 \times 504$ for OneVision-Encoder. This design avoids resolution-induced artifacts and enables a direct assessment of the encoder’s native-resolution modeling capability under realistic multimodal inference conditions.
\multimodalbenchmarktable

\paragraph{Codec-based Patch Sampling.}
We evaluate the same codec-guided patchification principle as described in Stage~2 of Section~\ref{subsec:pretraining-ove} under a strictly controlled token budget. For each test video, we uniformly sample $64$ frames over the full duration to obtain broad temporal coverage. We do not re-encode or transcode benchmark videos; compressed-domain signals (e.g., motion vectors and residuals) are extracted directly from the original bitstreams, thus preserving the native GOP structures and codec parameters of each dataset. Based on these signals, we compute a lightweight saliency score to estimate temporally informative regions and select patches sparsely across the sampled frames. The selected patches are then packed into a fixed number of visual tokens. Importantly, we keep the total token budget identical to the dense baseline that encodes only $8$ frames, ensuring that performance differences reflect improved token allocation rather than increased token count.

\paragraph{Comparison with SigLIP2.}
We first compare OneVision-Encoder with SigLIP2 under identical multimodal fine-tuning conditions, as shown in Table~\ref{tab:multimodal_benchmarks}. All models share the same instruction-tuning corpus, data preprocessing pipelines, training schedules, decoding strategies, and visual token budgets. Under this setting, OneVision-Encoder consistently outperforms SigLIP2 across 16 video, image, and document understanding benchmarks when integrated into an LMM built upon Qwen3-4B, indicating stronger multimodal transfer from the learned visual representations.

\paragraph{Comparison with Qwen3-ViT.}
We further conduct a comparison with Qwen3-ViT. Specifically, we integrate OneVision-Encoder with the Qwen3-1.7B language model and train it under the LLaVA-OneVision-1.5 framework, completing both Stage~1 and Stage~1.5 to adapt the encoder to native-resolution inputs. After this adaptation, the trained OneVision-Encoder is decoupled and compared with Qwen3-ViT under the same LLaVA-Next-Videos instruction-tuning setting. Under this setting, OneVision-Encoder outperforms Qwen3-ViT across 16 understanding benchmarks when evaluated with an LMM built upon Qwen3-4B, as shown in Table~\ref{tab:multimodal_benchmarks}. Notably, these gains are achieved despite OneVision-Encoder being pretrained on substantially fewer visual–text tokens (approximately 100B caption tokens), whereas Qwen3-ViT is pretrained on more than 2.1T caption and instruction-aligned tokens. This result suggests that the observed improvements arise from more effective visual representation learning, rather than increased pretraining scale or architectural specialization.

\subsubsection{Stage-wise Multimodal Training Analysis.}
We further analyze how stage-wise multimodal training contributes to the observed performance in LMM probing. 
In the first stage (OV-Encoder-Stage1), the visual encoder is trained using image-only data, focusing on general-purpose visual representation learning. 
In the second stage (OV-Encoder-Stage2), OCR and video data are introduced on top of Stage1, together with Codec patch selection and chunk-wise temporal sampling strategies.

As shown in Table~\ref{tab:ov_encoder_stage}, the Stage2 model consistently outperforms its Stage1 counterpart on multimodal and OCR-related benchmarks, while maintaining strong performance on general visual reasoning tasks. 
This comparison demonstrates that injecting OCR and video supervision plays a critical role in enhancing the ViT-based encoder’s suitability as a unified visual backbone for LMMs. 
Together with the native-resolution results above, this analysis highlights that stage-wise multimodal training is a key factor enabling both robust native-resolution generalization and effective multimodal reasoning.

\ovencoderstagetable

\subsection{Attentive Probing Evaluation}
\label{subsec:attentive_probe}

We evaluate the quality of visual representations learned by \ours using an attentive probing protocol, which has been widely adopted to assess backbone-level spatiotemporal modeling capacity without task-specific adaptation. In this setting, the visual encoder is frozen and a lightweight attention-based classifier head is trained on top of the extracted features for downstream video classification. This protocol isolates the intrinsic representational strength of the encoder and enables fair comparison across architectures with different tokenization and temporal modeling strategies.

\paragraph{Experimental Setup.}
All models are evaluated under a controlled and unified attentive probing protocol. 
Following prior work on vision-language pretraining and attentive pooling~\citep{siglip2}, we employ a multi-head attention pooling classifier to aggregate spatiotemporal features into video-level representations. 
The same classifier architecture is used for all methods to ensure architectural consistency, and the probing head is trained with an identical number of epochs, optimization settings, and learning rate schedules across all experiments.
All experiments are conducted on a cluster of 8 NVIDIA A800 GPUs. We therefore evaluate our model by
assessing the quality of the model’s learned representation on a set of seven benchmarks: SSV2, Diving48~\citep{li2018resound}, Perception Test~\citep{perceptiontest}, CharadesEgo~\citep{sigurdsson2018charades},  Epic-Kitchens-100 ~\citep{kay2017kinetics}, Kinetics-400~\citep{damen2022rescaling}, HMDB51~\citep{kuehne2011hmdb}. Batch sizes are determined on a per-dataset basis to balance computational efficiency and training stability: a batch size of 32 is used for SSV2, Diving48, and Perception Test, 16 for HMDB51, and 128 for all remaining datasets. During evaluation, we adopt a single-crop inference protocol, using one temporal crop and one spatial crop per video clip, resulting in a single prediction for each input video.
\videobenchmarktable
\textbf{}

\paragraph{Input Configuration.}
For frame-centric baselines, including SigLIP2, DINOv3, and AIMv2, we evaluate both 8-frame and 16-frame inputs using uniform temporal sampling. For \ours, we evaluate two instantiations under an identical patch budget of 512 visual tokens. Specifically, the codec-guided variant operates on dense 64-frame video inputs, where codec-derived signals determine the visible patch indices, while the chunk-wise variant samples frames within fixed temporal chunks. This design ensures that all methods are compared under a fixed token budget, decoupling representational capacity from raw input resolution or frame count.

\patchscalingtable

\paragraph{Evaluation Protocol.}
During inference, all crops belonging to the same video are aggregated by averaging logits across crops. Labels are shared across crops of the same video, and no test-time augmentation beyond the single-crop setting is applied. For codec-based models, visible patch indices are provided as part of the input to ensure consistent patch selection across crops.

\paragraph{Results.}
Table~\ref{tab:video_benchmarks} shows that \ours consistently outperforms SigLIP2, DINOv3, and AIMv2 across all evaluated video benchmarks under both 8-frame and 16-frame settings. In particular, \ours achieves a \textbf{12.7\% absolute Top-1 accuracy improvement} over SigLIP2 on Diving48 at an identical patch budget, demonstrating superior motion modeling capability. Notably, these gains are achieved without sacrificing performance on appearance-dominated datasets such as Kinetics-400, confirming that Codec patch-level encoding yields more discriminative and efficient spatiotemporal representations. We further observe that for 64-frame inputs under a fixed token budget, allocating only one to two I-frames is sufficient to establish stable spatial context, with subsequent frames contributing primarily through sparse motion-driven updates.

\subsection{Patch-Efficient Video Understanding Comparison}

We conduct an efficiency analysis comparing SigLIP2 with dense full-frame patch processing and \ourscodec under a fixed token budget, as shown in Table~\ref{tab:patch_scaling}. It is important to emphasize that \ourscodec does not perform temporal downsampling of the input video. All results are obtained from the same 64-frame (16384 patches) source video, where codec-native motion vectors and residuals are used to selectively extract a fixed number of spatiotemporal patches distributed across the entire temporal extent.

For a fair comparison, SigLIP2 is evaluated under the same token budgets and adopts a traditional frame sampling strategy, where each group of 256 patches corresponds to a contiguous RGB frame. Under a fixed token budget, \ourscodec redistributes patches across time while preserving their spatial positions, enabling long-range temporal coverage. As a result, it outperforms SigLIP2 on Diving48 and Perception Test while reducing patch processing by 75.0\%–96.9\% compared to dense processing of 16,384 patches. Specifically, the reduction ratio is computed relative to the dense baseline that processes all $64 \times 256 = 16{,}384$ patches from the full video. Using token budgets of 4096, 2048, 1024, and 512 patches corresponds to retaining 25.0\%, 12.5\%, 6.25\%, and 3.1\% of the dense patches, respectively, yielding a patch reduction of 75.0\%–96.9\%.

\codecselectedmotion

\subsection{Ablation of Codec-guided Patch Selection}
\label{subsec:causal_codec}

Although \ourscodec consistently outperforms frame-centric baselines under attentive probing, as shown in Table~\ref{tab:video_benchmarks}, performance gains alone do not establish codec-guided patch selection as a functional mechanism. To isolate its causal role, we conduct a set of controlled interventions that explicitly manipulate patch content while holding token count, spatiotemporal positions, positional encodings, and model parameters fixed.

\paragraph{Patch Content Necessity.}
We first examine whether the content of codec-selected motion patches is necessary for the observed gains. In this setting, motion-heavy patches identified by the codec are replaced with non-motion patches sampled from the same video, while preserving their original spatiotemporal positions. As shown in Table~\ref{tab:causal_interventions}, this intervention leads to substantial performance degradation across all benchmarks. The drop is particularly pronounced on motion-sensitive datasets, with accuracy decreasing, while appearance-dominated datasets such as Kinetics-400 exhibit smaller but consistent declines. These results indicate that the benefits of \ourscodec cannot be attributed to token sparsity or positional bias alone, but critically depend on the motion-centric content encoded by the selected patches.

\begin{figure*}
    \centering
    \includegraphics[width=1\linewidth]{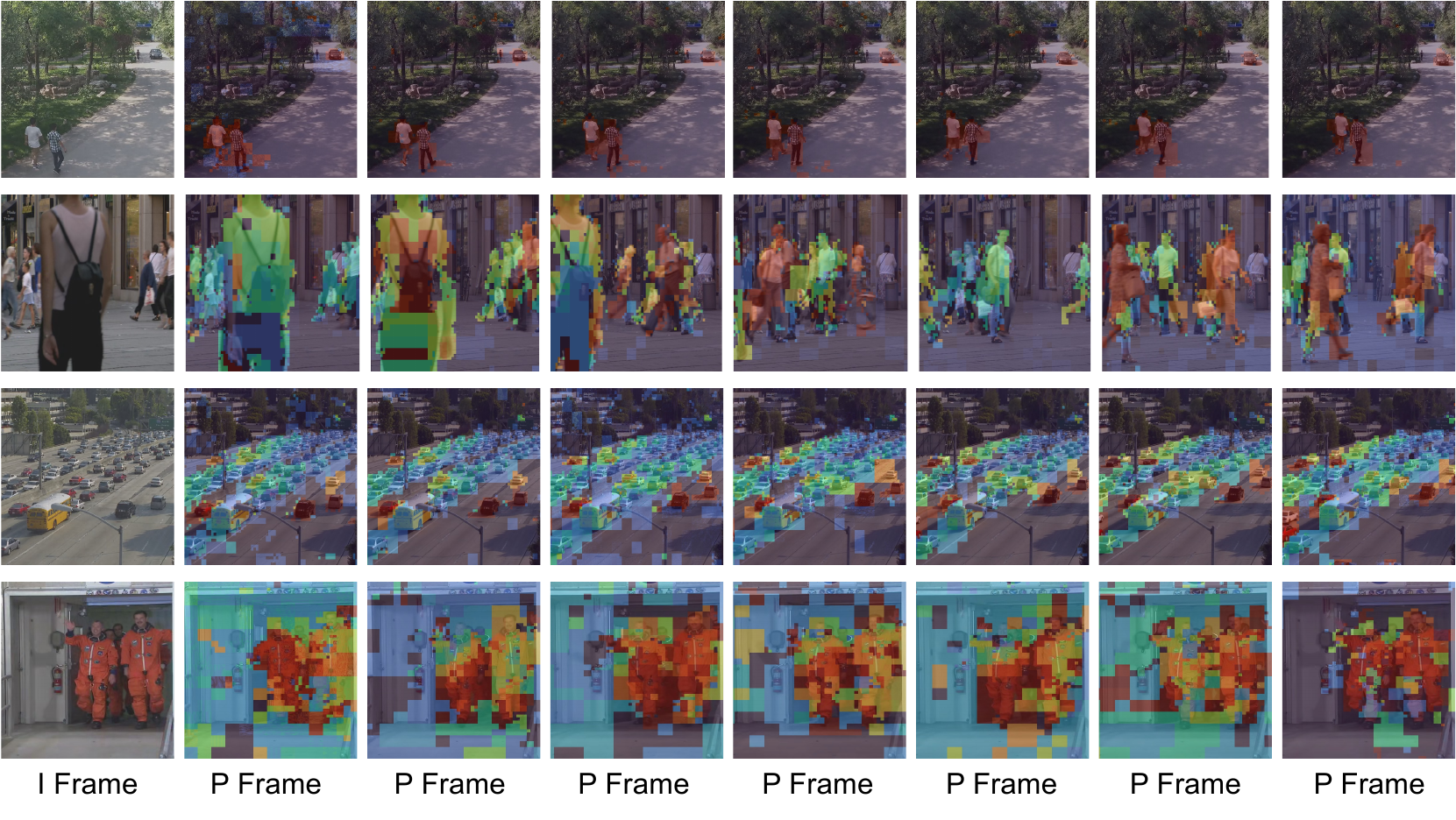}
    \caption{Visualization of I- and P-frame decomposition in HEVC. I-frames retain complete spatial structure, whereas P-frames encode motion-compensated residuals highlighting motion. Bright areas denote high residual magnitudes, while dark areas indicate static content.}
    \label{fig:pca_local}
\end{figure*}

\paragraph{Semantic Specificity of Motion Cues.}
To further assess whether the model relies on semantically aligned motion rather than generic motion signals, we perform a counterfactual replacement in which codec-selected motion patches are substituted with motion patches drawn from unrelated videos. Despite preserving motion magnitude and patch positions, this intervention results in even larger performance drops on fine-grained temporal benchmarks, as shown in Table~\ref{tab:causal_interventions}. The consistent degradation relative to non-motion replacement demonstrates that \ourscodec is sensitive to the semantic correctness of motion cues, rather than merely benefiting from the presence of motion energy or stochastic perturbations.

\paragraph{Negative Control: Patch-Position Shuffle.}
As a sanity check, we additionally evaluate a patch-position shuffle intervention, in which the visual content of codec-selected patches is preserved but their spatiotemporal positions are randomly permuted. This intervention causes a substantially larger performance drop across all benchmarks, as shown in Table~\ref{tab:causal_interventions}, confirming that coherent spatial and temporal alignment is critical for effective representation learning. Importantly, this experiment is not intended to demonstrate the causal role of codec guidance itself, but rather to rule out degenerate explanations in which patch content alone suffices without positional structure.

\paragraph{Discussion.}
Taken together, these interventions establish a consistent ordering across benchmarks: preserving semantically correct motion patches yields the strongest performance, followed by non-motion substitutions, while semantically mismatched motion patches and patch-position shuffling are most detrimental. This hierarchy rules out alternative explanations based on regularization, noise injection, or attention disruption, and provides converging evidence that codec-guided patch selection captures motion-centric visual evidence that is both semantically meaningful and structurally aligned with the underlying video content.

\subsection{Qualitative Analysis}
To understand how our model leverages temporal information for selective patch processing, we visualize the patch selection mechanism guided by residual magnitudes and motion vectors. As illustrated in Figure~\ref{fig:pca_local}, we compare an I-frame (reference frame) with subsequent P-frames in a video sequence, where patch colors indicate their selection priority based on residual and motion strength. In the I-frame (top-left), all patches are processed uniformly since no temporal prior exists. In contrast, the P-frames show selective emphasis on patches with large residuals and strong motion vectors, corresponding to regions with significant appearance changes or object movements. The highlighted patches, shown in warmer colors  (\textcolor{red}{red}, \textcolor{orange}{orange}, \textcolor{yellow}{yellow}), primarily correspond to dynamic foreground objects such as moving pedestrians, whereas static background regions (trees, buildings) appear in \textcolor{blue}{cooler colors} (\textcolor{blue}{blue}, \textcolor{green}{green}), indicating reduced computational focus. Across the sequence, the model consistently tracks salient motion regions, as pedestrians maintain high activation throughout their trajectories, demonstrating that the residual–motion criterion effectively identifies temporally informative patches. This visualization confirms that our approach achieves spatial selectivity by concentrating computation on motion-rich areas and allocating representational capacity according to temporal saliency, leading to more efficient video understanding.

\begin{figure*}[!t]
    \centering
    \includegraphics[width=\linewidth]{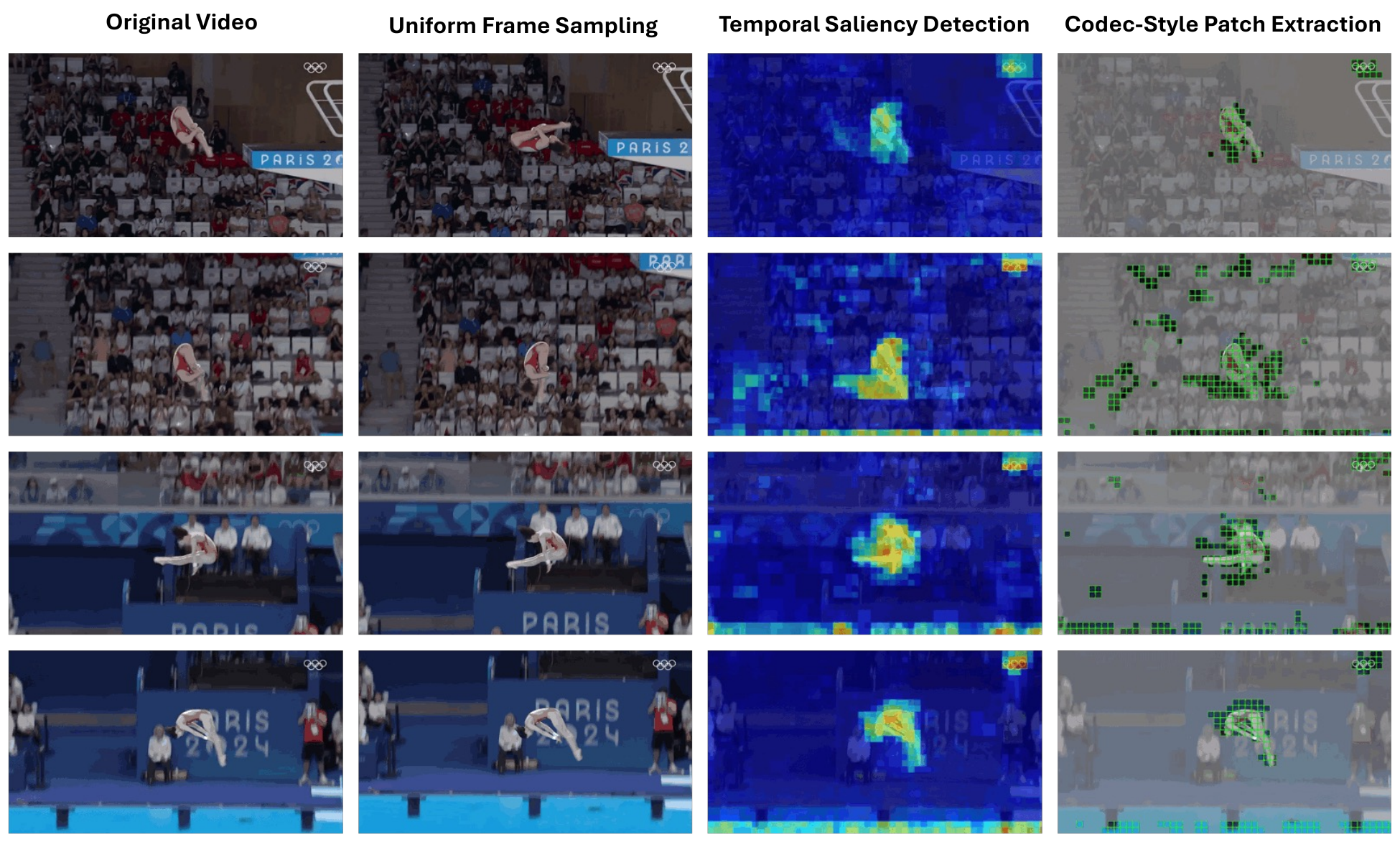}
    \caption{Comparison of video processing pipelines for spatiotemporal representation learning. (a) original dense video input with full temporal context, (b) uniform frame sampling that sparsely selects evenly spaced frames, (c) temporal saliency detection that identifies motion- and event-centric regions across all frames, and (d) Codec patch extraction that selectively retains temporally salient patches under a fixed token budget.}
\label{fig:case1}
\end{figure*}

Figure~\ref{fig:case1} contrasts conventional frame-centric video processing with the proposed Codec patch extraction. While dense 64-frame inputs preserve full temporal context at high computational cost, uniform frame sampling reduces computation by sparsely selecting frames but inevitably discards fine-grained inter-frame motion, particularly for fast or subtle actions. Temporal saliency detection instead analyzes all frames to identify motion- and event-centric regions. Leveraging this signal, Codec patch extraction selectively encodes only temporally salient patches using a codec-inspired ordering, achieving substantial token reduction (75\%–96.9\%) while preserving critical motion dynamics. This formulation decouples temporal coverage from token density, enabling efficient and scalable spatiotemporal modeling without reliance on sparse frame sampling.

\section{Related Work}

\subsection{Advances in Visual Representation Learning}
Vision Transformers~\citep{dosovitskiy2021image,li2022uniformer} have emerged as a dominant approach in visual representation learning. DINOv2~\citep{dinotext} and DINOv3~\citep{simeoni2025dinov3} introduce a self-supervised framework that does not require labels, achieving state-of-the-art results across a wide range of vision tasks. Simultaneously, equivariant self-supervised methods~\citep{devillers2022equimod,park_learning_2022,garrido2023sie,gupta2023care,dangovski_equivariant_2022} have been developed to predict structured transformations consistent with group-theoretic principles. Masked image modeling techniques~\citep{he2021mae,bao2021beit,el2024aim,xie2022simmim} learn visual representations by reconstructing masked regions within the pixel domain. Furthermore, Joint-Embedding Predictive Architecture~\citep{assran2023ijepa,assran2025v,baevski2022data2vec} showed that predicting a learned latent space instead of the pixel space leads to more powerful, higher-level features. Contrastive Language-Image Pretraining (CLIP)~\citep{bolya2025perception,EVA-CLIP,li2023clipav2,dfn,metaclip,laion} aligns images and texts within a shared embedding space through instance-level contrastive supervision. However, existing methods mainly focus on either pixel-level reconstruction or instance-level contrastive alignment, limiting their ability to capture structured semantic relationships across samples and modalities. In this work, we adopt the self-supervised cluster discrimination paradigm (\textit{e.g.}, UNICOM~\citep{an2023unicom}, MLCD~\citep{an2024multi}, RICE~\citep{xie2025region}, UniViT~\citep{tang2025univit}), which learns structured semantics by jointly clustering similar instances and discriminating between clusters.

\subsection{Efficient Video Representations} 

\textbf{(1) Video Sampling.} Conventional methods first divide selected frames into fixed patch embeddings (\textit{e.g.}, 16×16 in ViT) before encoding, where frame selection typically follows uniform sampling or motion-based heuristics~\citep{xu2023hdvila,li2023internvid}. However, such strategies still process all patches per frame, resulting in long token sequences and high preprocessing cost. To improve efficiency, patch-level merging is adopted in AuroraCap~\citep{chai2024auroracap} and AuroraLong~\citep{xu2025auroralong}, while adaptive sampling~\citep{kim2024image} allocates more frames to motion-intensive segments but fails to mitigate the quadratic cost of self-attention under large token budgets. Moreover, flexible-FPS variants impose an upper bound on frame count, limiting scalability to high frame rates. Video-LaVIT~\citep{jin2024video} explores decomposing videos into keyframes and motion signals to enable efficient tokenization and unified generative pretraining with multimodal LLMs. \textbf{(2) Token Dropout.} A simple strategy to reduce transformer sequence length is token dropout, which removes redundant tokens either randomly~\citep{han2022turbo,liu2023patchdropout} or through learned selection~\citep{rao2021dynamicvit,yin2022vit,chen2023efficient}. PatchDropout~\citep{liu2023patchdropout} applies random token removal to standard ViTs for faster training while retaining all tokens at inference, whereas Turbo~\citep{han2022turbo} adopts partial masking for video reconstruction. However, token dropout often removes informative regions and disrupts temporal consistency, leading to unstable optimization and weakened motion modeling. \textbf{(3) Token Merging.} Various approaches have been employed to merge tokens in order to mitigate the information loss associated with dropout. In~\citep{liang2022not}, inattentive tokens are combined into a singular "background" token, while other studies utilize semantics to enhance the features of interest~\citep{zeng2022not,zhou2023can}. Additional research~\citep{lee2024video} focuses on optimizing token merging through the use of trainable parameters. The latest work by~\citep{koner2024lookupvit} trains models using an additional stream of compressed tokens to minimize the overhead of attention. Alternatively, numerous studies have implemented token merging, as detailed in~\citep{bolya2022token,bolya2023token}, where image tokens are merged via a weighted average, leveraging attention keys as a similarity metric. In this work, we do not aim to improve compression efficiency or replace existing codecs. Instead, we treat the structural decomposition exposed by modern video codecs as a guiding principle for visual representation learning, where stable spatial context is separated from sparse, motion driven variations. This perspective directly informs the design of our spatiotemporal transformer.

\subsection{Video Codec Compression}
Video codec compression has been extensively studied, with a large body of work devoted to designing efficient and effective video coding systems~\citep{1369699, 4317622, kumar2019novel}. Classical standards such as H.264/AVC~\citep{sullivan2012overview, zhao2006highly} established the foundation of modern video compression by introducing key techniques including motion compensation, transform coding, and entropy coding, leading to substantial gains in compression efficiency. Building upon this framework, the High Efficiency Video Coding (HEVC) standard, also known as H.265~\citep{sullivan2005video,sullivan2012overview}, further improved coding performance through more expressive block partitioning, enhanced motion modeling, and refined entropy coding mechanisms.

Beyond traditional hand-crafted codecs, recent research has explored deep learning-based video compression methods~\citep{li2021deep,mentzer2022vct, 9288876, umt, zhang2023extracting, wang2025make}, which employ neural networks to model spatial and temporal redundancies directly from data. These approaches have demonstrated promising compression performance and have also motivated perceptually driven coding strategies that allocate bits according to human visual sensitivity. Efficiency has also been explored in transformer-based video models, where methods such as Run-Length Tokenization (RLT)~\citep{choudhury2024don} exploit temporal redundancy at the token level to reduce input tokens while maintaining performance. Recent work such as EMA~\citep{zhao2025efficient} further leverages compressed video streams by encoding GOP structures with motion-aware mechanisms for efficient video MLLM understanding. Collectively, prior work in video compression has made significant progress toward compactly representing video content by exploiting its inherent temporal predictability and structural redundancy. In this work, we do not aim to improve compression efficiency or replace existing codecs; instead, we draw inspiration from the structural principles underlying modern video codecs, particularly their explicit decomposition of spatial context and temporal variation, to guide the design of spatiotemporal representations for visual transformers.

\section{Conclusion}

In this work, we introduced \ours, a unified self-supervised vision framework that departs from frame-centric modeling and aligns representation learning with the predictive structure of video signals. By treating discriminative visual evidence as patch-level and motion-centric, OneVision-Encoder selectively encodes informative regions while preserving dense temporal coverage under fixed token budgets. Central to this design is Codec Patchification, which constructs sparse yet structure-preserving spatiotemporal token layouts and naturally extends to chunk-wise temporal modeling and single-image inputs within a unified attention-based encoder, supported by 3D rotary positional encoding. Combined with a cluster discrimination objective that jointly models object-level and motion-level semantics without external supervision, OneVision-Encoder achieves state-of-the-art performance under both LMM probing and attentive probing. These results highlight codec-inspired patch-level sparsity as an effective and scalable foundation for general-purpose visual representation learning.

\section{Contributors}

\begin{minipage}[t]{0.48\textwidth}
\raggedright
\textbf{Contributors} \\ 
{\color{gray}\small core contributors are in bold}

\begin{itemize}[noitemsep,topsep=0pt,leftmargin=*]
\item \textbf{Feilong Tang}
\item \textbf{Xiang An}
\item \textbf{Yunyao Yan}
\item \textbf{Yin Xie}
\item Bin Qin
\item Kaicheng Yang
\item Yifei Shen
\item Yuanhan Zhang
\item Chunyuan Li
\item Shikun Feng
\item Changrui Chen
\item Huajie Tan
\item Ming Hu
\item Manyuan Zhang

\end{itemize}
\end{minipage}\hfill
\begin{minipage}[t]{0.48\textwidth}
\raggedright
\textbf{Project Leaders}

\begin{itemize}[noitemsep,topsep=0pt,leftmargin=*]
\item Bo Li
\item Ziyong Feng
\item Ziwei Liu
\item Zongyuan Ge
\item Jiankang Deng
\end{itemize}
\end{minipage}

\clearpage

\section{Implementation Details}
\label{sec:implementation}

\paragraph{Model Architecture and Configuration.}
OneVision-Encoder Large is implemented as a Vision Transformer with 24 transformer layers, a hidden dimension of 1024, and 16 attention heads. The model uses a patch size of $14 \times 14$ and adopts GELU activations with Layer Normalization throughout the network. Attention computation is accelerated using Flash Attention 2, enabling efficient training and inference at scale. A summary of the core architectural hyperparameters is provided in Table~\ref{tab:arch_details}.

\begin{table}[!h]
\centering
\small
\caption{Architecture configuration of OneVision-Encoder Large.}
\label{tab:arch_details}
\begin{tabular}{lc}
\toprule
\textbf{Component} & \textbf{Setting} \\
\midrule
Transformer layers & 24 \\
Hidden dimension & 1024 \\
Attention heads & 16 \\
Patch size & $14 \times 14$ \\
MLP expansion ratio & 4$\times$ \\
Position encoding & 3D RoPE (T:H:W = 4:6:6) \\
\bottomrule
\end{tabular}
\end{table}

\paragraph{Unified Patch-based Input Representation.}
All inputs are converted into patch tokens and processed by a single ViT backbone. Images are treated as single-frame videos ($T{=}1$), while videos are represented in a 5D tensor format. To ensure consistent temporal reasoning, all video inputs are mapped to a virtual temporal grid of 64 frames, regardless of the actual number of frames processed. This mapping enables uniform temporal position encoding across dense, sparse, and Codec inputs.

For inputs that do not cover all 64 frames explicitly, a \emph{visible indices} mechanism is used to associate each selected patch with its corresponding temporal position in the virtual grid. This design decouples temporal coverage from token density and allows sparse inputs to preserve long-range temporal structure.

\paragraph{Codec-style Patch Selection.}
Codec-style processing operates on dense videos of 64 frames at a spatial resolution of $224 \times 224$. Motion vectors and prediction residuals are extracted from the HEVC codec and used to estimate patch-level temporal saliency. Motion vectors capture object displacement at sub-pixel precision, while residuals encode fine-grained appearance changes not explained by motion compensation. These signals are fused into a unified saliency score for each patch across all frames.

Patches are ranked globally by their saliency scores, and only the top-$K$ patches are retained. This selection typically preserves between 3.1\% and 25\% of all patches, corresponding to a compression ratio of 75\%–96.9\% relative to dense processing. Selected patches are then reassembled into a compact video representation and passed to the ViT using sparse visible indices. Table~\ref{tab:codec_selection} summarizes the codec-style selection procedure.

\begin{table}[!h]
\centering
\caption{Codec-style patch selection pipeline.}
\label{tab:codec_selection}
\begin{tabular}{lp{10cm}}
\toprule
\textbf{Step} & \textbf{Description} \\
\midrule
Motion extraction & Decode HEVC motion vectors with camera motion compensation \\
Residual extraction & Obtain prediction residuals for fine-grained changes \\
Energy fusion & Combine motion and residual energies into a saliency score \\
Top-$K$ selection & Retain globally most salient patches across all frames \\
Sparse encoding & Process selected patches with sparse visible indices \\
\bottomrule
\end{tabular}
\end{table}

\paragraph{Video Processing Modes and Batch Composition.}
During pretraining, OneVision-Encoder employs a mixed-modality batch that includes multiple video processing modes. This design exposes the model to diverse temporal structures and encourages robust representation learning. Video samples within a batch are split into three processing modes: Codec patchification, uniform frame sampling, and Tiling-style spatial concatenation. All modes produce inputs that are compatible with the same ViT backbone and position encoding.

\begin{table}[t]
\centering
\small
\caption{Video processing modes used during training.}
\label{tab:video_modes}
\begin{tabular}{lccc}
\toprule
\textbf{Mode} & \textbf{Batch Ratio} & \textbf{Input Form} & \textbf{Output Shape} \\
\midrule
Codec & 50\% & Dense video + saliency & $[B, 3, 8, 224, 224]$ \\
Frame sampling & 37.5\% & Uniform temporal bins & $[B, 3, 8, 224, 224]$ \\
Tiling & 12.5\% & Vertical frame concatenation & $[B, 3, 1792, 224]$ \\
\bottomrule
\end{tabular}
\end{table}

\paragraph{Position Encoding Consistency.}
All input modes share the same 3D Rotary Position Embedding. The temporal dimension of the RoPE encodes the position of each patch within the 64-frame virtual grid, while spatial dimensions encode patch row and column indices. For Codec inputs, patches may originate from arbitrary frames but are positioned correctly via their temporal indices. For uniformly sampled frames, temporal gaps are explicitly encoded. Tiling inputs are treated as single-frame inputs with fixed temporal positions. This unified encoding scheme enables the model to reason coherently over heterogeneous spatiotemporal layouts.

\paragraph{Training and Inference Behavior.}
The model is trained using a unified optimization pipeline across all modalities. No modality-specific parameters or task-specific encoders are introduced. At inference time, the same preprocessing logic is applied, and the model can flexibly switch between Codec sparse processing and conventional frame sampling depending on computational constraints. This design allows efficient deployment across a wide range of image and video understanding tasks without architectural modification.

\section{Controlled Evaluation Pipeline}
\begin{figure}[h]
\centering
\includegraphics[width=\linewidth]{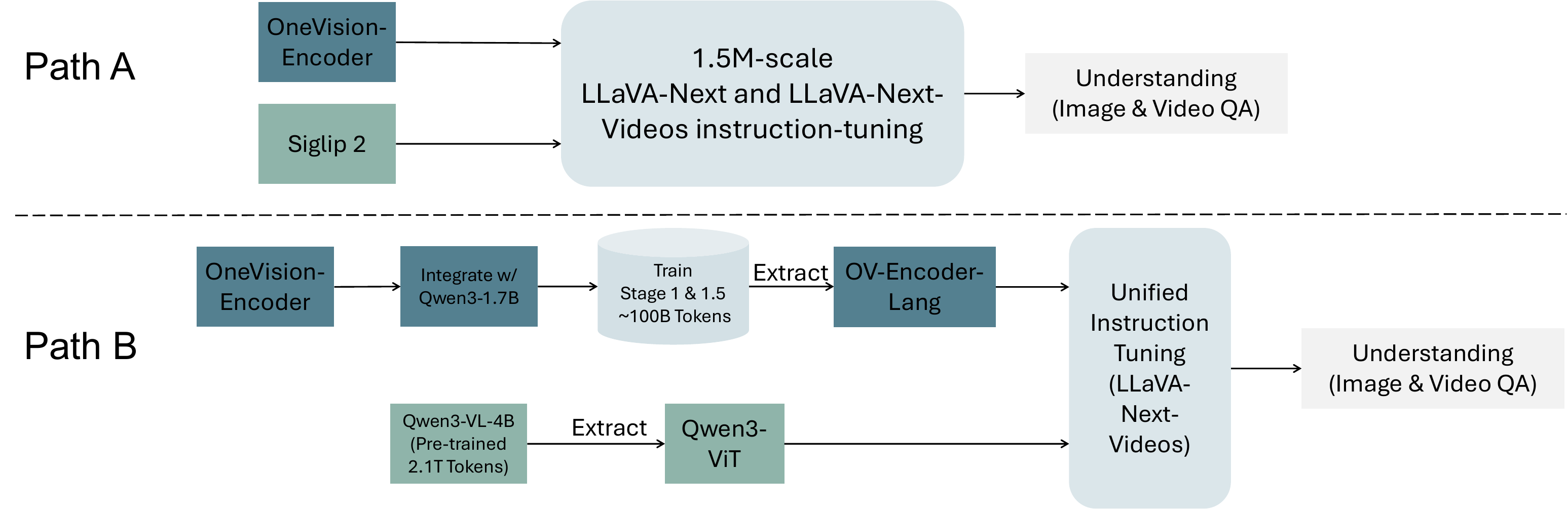}
\caption{Controlled evaluation pipeline decoupling the encoder for fair comparison against Qwen3-ViT and SigLIP2}
\label{fig:LLMProbing}
\end{figure}

As shown in Figure~\ref{fig:LLMProbing}, we adopt a controlled evaluation protocol to compare OneVision-Encoder with Qwen3-ViT and SigLIP2 under LMM probing. For comparison with Qwen3-ViT, OneVision-Encoder is first integrated with the Qwen3-1.7B language model and trained through Stage 1 and Stage 1.5 under the LLaVA-OneVision-1.5 framework to adapt the encoder to native-resolution multimodal inputs. After alignment, the trained vision encoder is decoupled and evaluated under the same LLaVA-Next-Videos instruction-tuning setting as Qwen3-ViT, ensuring a fair comparison under identical downstream supervision. For comparison with SigLIP2, all models are directly evaluated under identical multimodal fine-tuning conditions using a unified 1.5M-scale instruction-tuning corpus, while keeping the language model backbone fixed. This decoupled and unified evaluation pipeline isolates the contribution of visual representation learning and avoids confounding effects from language model capacity, instruction data leakage, or differing alignment procedures.

\begin{figure}[!t]
\centering
    \includegraphics[width=0.7\linewidth]{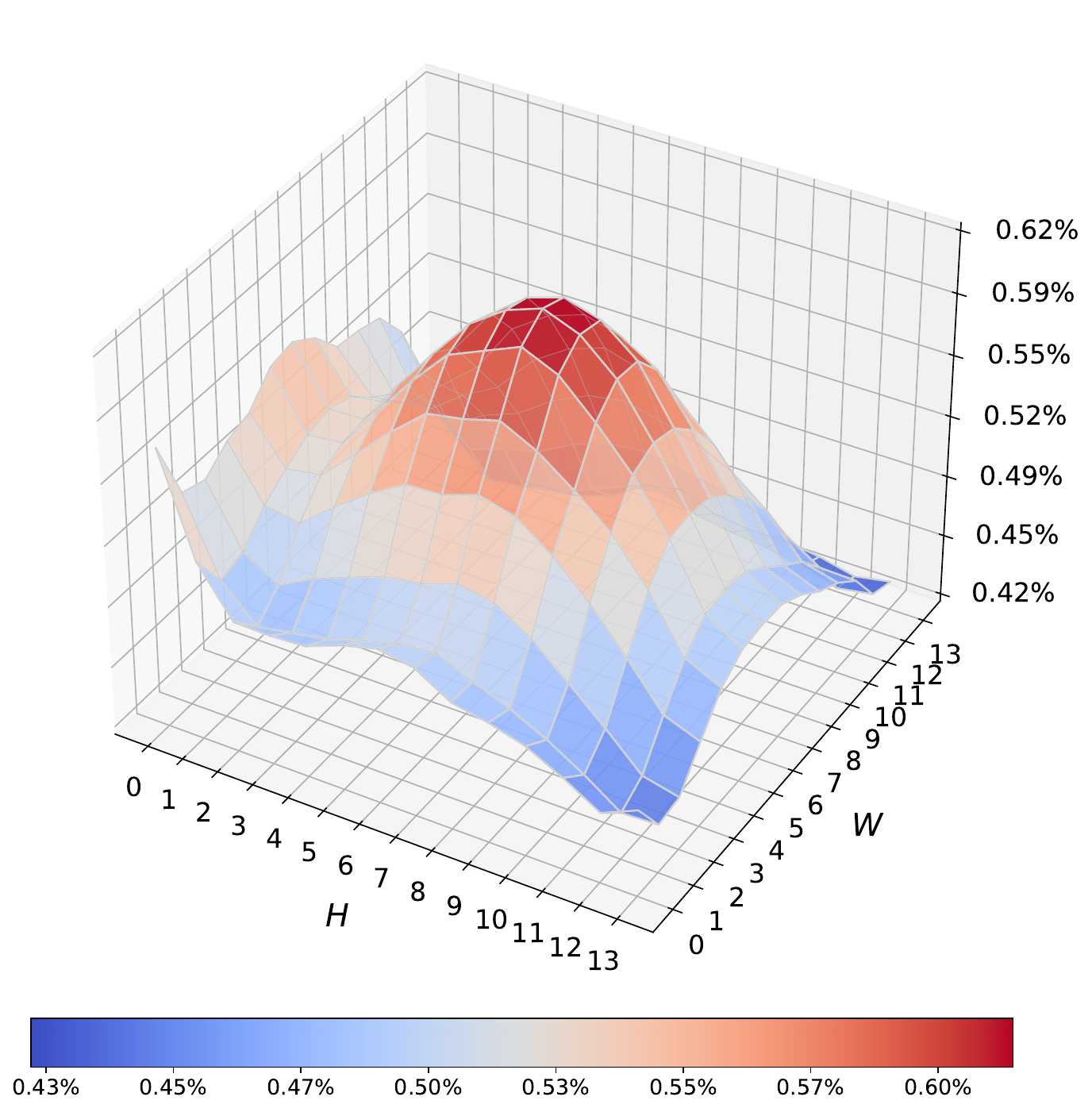}
    \caption{Spatial Bias Analysis of Selected Visual Patches under a given setting.}
    \label{fig:tokens}
\end{figure}

\section{Spatial Bias Analysis} 

As shown in Figure~\ref{fig:tokens}, the reported statistics are computed from a random subset of 200,000 video samples drawn from the training data. When using codec-guided patch selection alone, the selected visual tokens exhibit a pronounced spatial center bias, with the majority of tokens concentrated in the central regions of the frame. This behavior reflects intrinsic statistical properties of video data: due to camera framing and subject placement, salient motion cues and residual signals are typically denser near the image center. While such a bias enables the model to focus on regions with strong motion signals, it also leads to insufficient coverage of peripheral areas, thereby weakening the representation of global scene structure and fine-grained action cues.

With the introduction of chunk-wise patchification, the spatial distribution of selected tokens becomes markedly more uniform, redistributing tokens toward the image periphery and boundary regions and effectively mitigating the center bias induced by codec-driven selection. This effect arises because chunk-wise sampling partitions the video along the temporal dimension and performs global selection over visible patch indices, resulting in a more structurally balanced spatial coverage. Importantly, this rebalancing is achieved without increasing the token budget, but through a principled reallocation of patch selection that enhances spatial diversity and complements motion-centric evidence with global contextual information.

\section{Token Allocation Case Study}

For each case, we plot time (x-axis) versus accumulated visual tokens (y-axis) under a fixed token budget. Stepwise growth indicates when a method allocates a non-trivial number of tokens to a timestamp. Vertical rugs at the bottom show the actual timestamps observed by each strategy, and triangles denote I-frame anchors in the codec-style layout. Case~1 uses a 2048-token budget (uniform 8-frame baseline), and Case~2 uses a 4096-token budget (uniform 16-frame baseline). Detailed experiments and evaluation protocols for these two motion regimes will be reported in LLaVA-OneVision~2.

\subsection{Case 1}
\label{subsec:case1_continuous}

\begin{figure*}[!t]
    \centering
    \includegraphics[width=\linewidth]{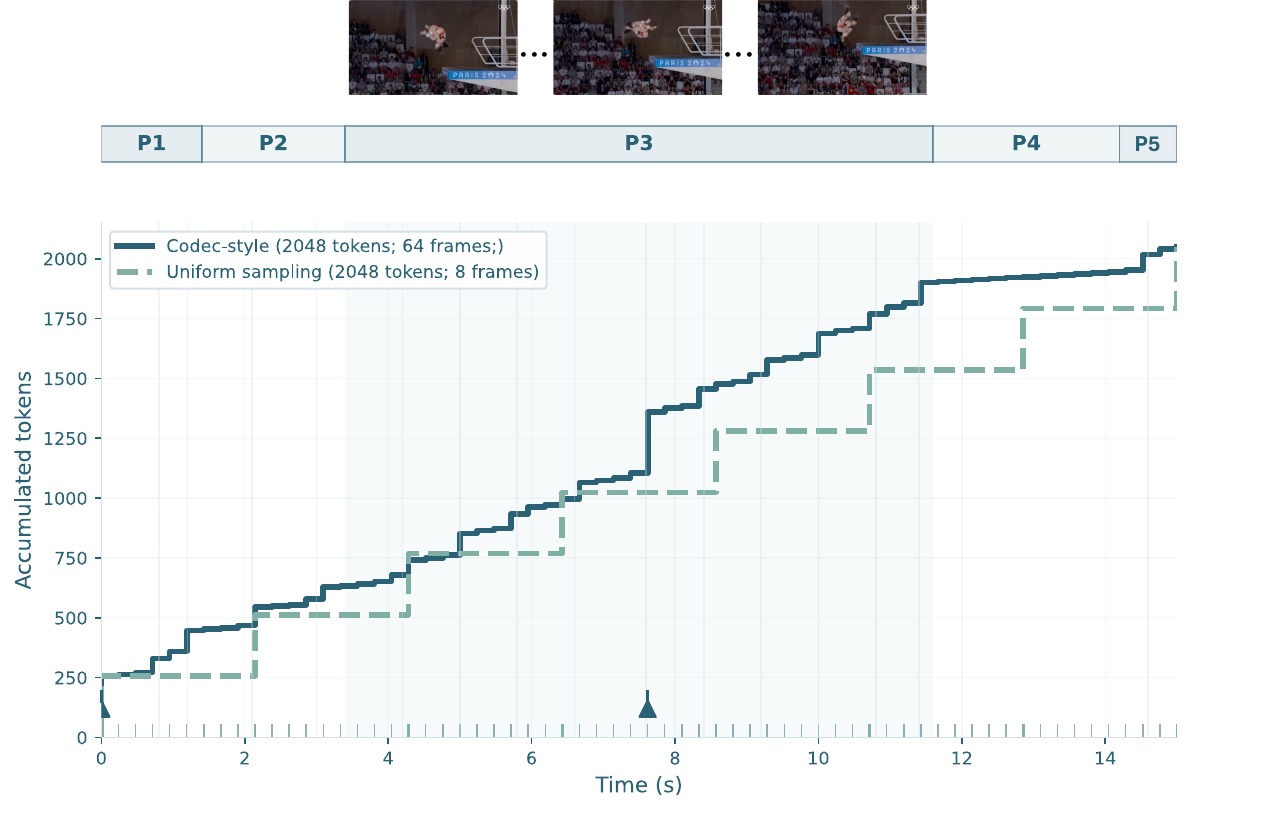}
    \caption{\textbf{Case study 1 (Diving): dense evidence for continuous motion.} Under a fixed budget of 2048 tokens, codec-style patch extraction distributes tokens over 64 sampled frames to capture dense pose transitions, while uniform sampling allocates the same budget to only 8 frames and may miss brief but critical instants. Phases: \textbf{P1} 0.0--1.4s (takeoff + angular momentum), \textbf{P2} 1.4--3.4s (tuck/pike set; body shape), \textbf{P3} 3.4--11.6s (twist+somersault progression; dense cues), \textbf{P4} 11.6--14.2s (spot water + open/align), \textbf{P5} 14.2--15.0s (entry alignment; line/splash). Diving is high-speed and continuous, so discriminative cues are temporally dense and benefit from dense coverage.}
    \label{fig:case_study1}
\end{figure*}

We consider videos whose semantics are carried by smooth, persistent motion, where essentially every moment contributes evidence. In this regime, missing intermediate timestamps breaks trajectory continuity and harms recognition. Figure~\ref{fig:case_study1} visualizes this behavior on a Diving example. Under the same fixed token budget of 2048 tokens, uniform \emph{8-frame} sampling concentrates all tokens on only eight sparse timestamps (leaving large temporal gaps), whereas codec-style allocation first samples a 64-frame timeline and then reallocates tokens through saliency-guided P-frame patch selection, preserving dense motion evidence across time. This is particularly important for high-angular-velocity actions such as diving, where discriminative cues are temporally dense and distributed across many brief pose transitions rather than confined to a single transition.

\subsection{Case 2}
\label{subsec:case2_instantaneous}

\begin{figure*}[!t]
    \centering
    \includegraphics[width=\linewidth]{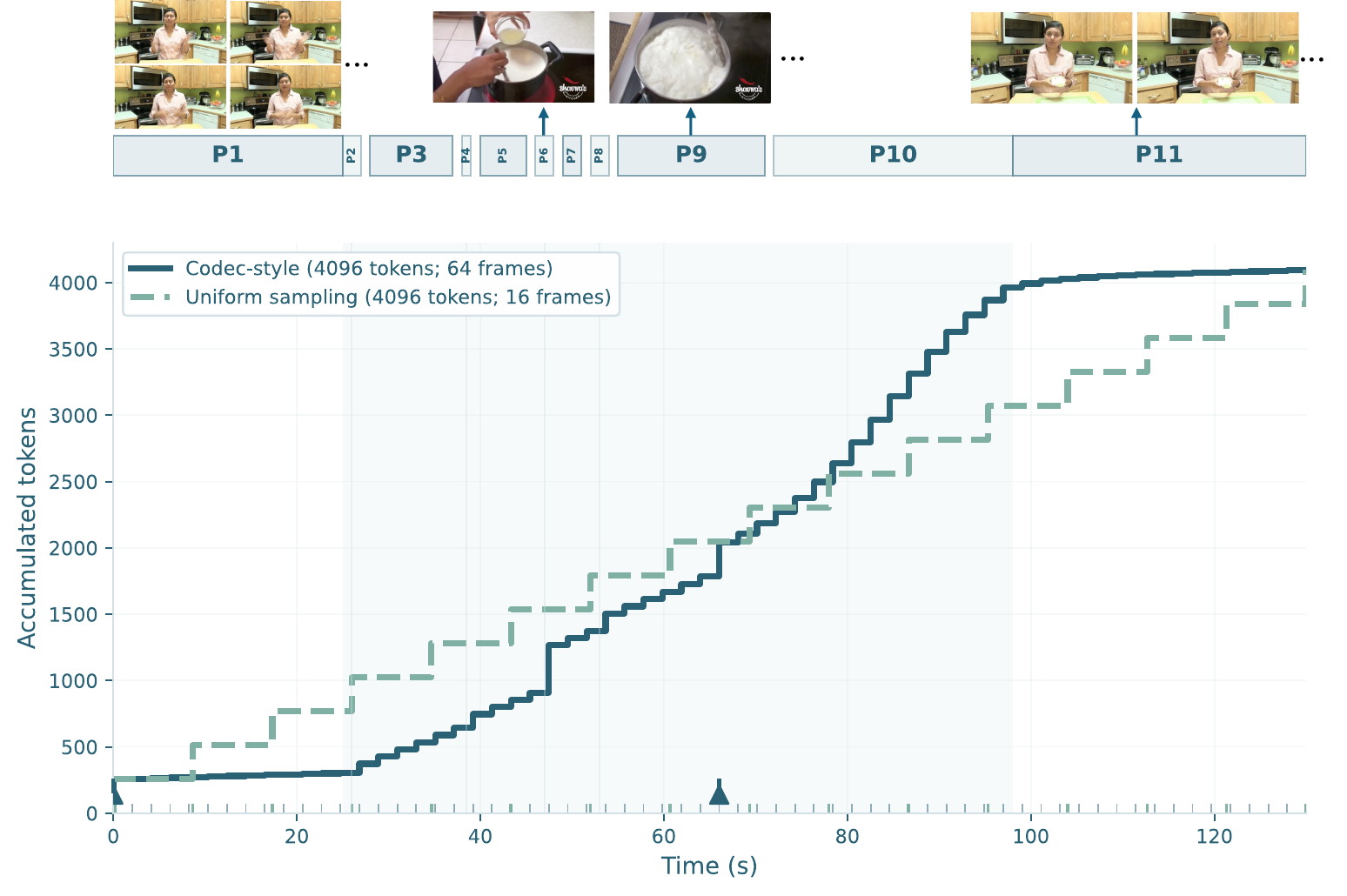}
    \caption{\textbf{Case study 2 (Cooking): sparse key frames under instantaneous motion.}
    We analyze a 130-second cooking video where decisive evidence concentrates in short, discrete transitions (e.g., brief pours) rather than being uniformly distributed over time. Under a fixed budget of 4096 tokens, codec-style allocation spreads tokens over a 64-frame timeline and reallocates P-frame tokens toward high-saliency moments, while uniform sampling allocates the same budget to only 16 frames and may miss brief segments entirely (misplacement). Phases: \textbf{P1} 0.0--25.0s (context), \textbf{P2} 25.0--27.0s (pour raw materials), \textbf{P3} 28.0--37.0s (mix raw materials), \textbf{P4} 38.0--39.0s (pour raw materials), \textbf{P5} 40.0--45.0s (mix raw materials), \textbf{P6} 46.0--48.0s (pour raw materials), \textbf{P7} 49.0--51.0s (mix raw materials), \textbf{P8} 52.0--54.0s (pour raw materials), \textbf{P9} 55.0--71.0s (mix raw materials), \textbf{P10} 72.0--98.0s (pour raw materials), \textbf{P11} 98.0--130.0s (context).}
    \label{fig:case_study2}
\end{figure*}

We consider videos whose semantics concentrate in short, discrete transitions, where the true ``key frames'' are sparse events such as brief pours, quick tool interactions, or sudden scene changes. In this regime, the primary risk of uniform sampling is \emph{misplacement}: even with a sufficient token budget, the 16 sampled frames can easily miss a short segment entirely, causing decisive evidence to be absent from the visual input. Figure~\ref{fig:case_study2} illustrates this effect on a 130-second cooking video. Notably, several pour segments are extremely short (e.g., \textbf{P4}: 38.0--39.0s), making them easy to skip under uniform sampling. In contrast, codec-style allocation first samples a 64-frame timeline and then reallocates the same token budget toward high-saliency moments via P-frame patch selection, increasing the probability of capturing the decisive evidence that later multi-turn QA may depend on.

\begin{figure*}
    \centering
    \includegraphics[width=0.95\linewidth]{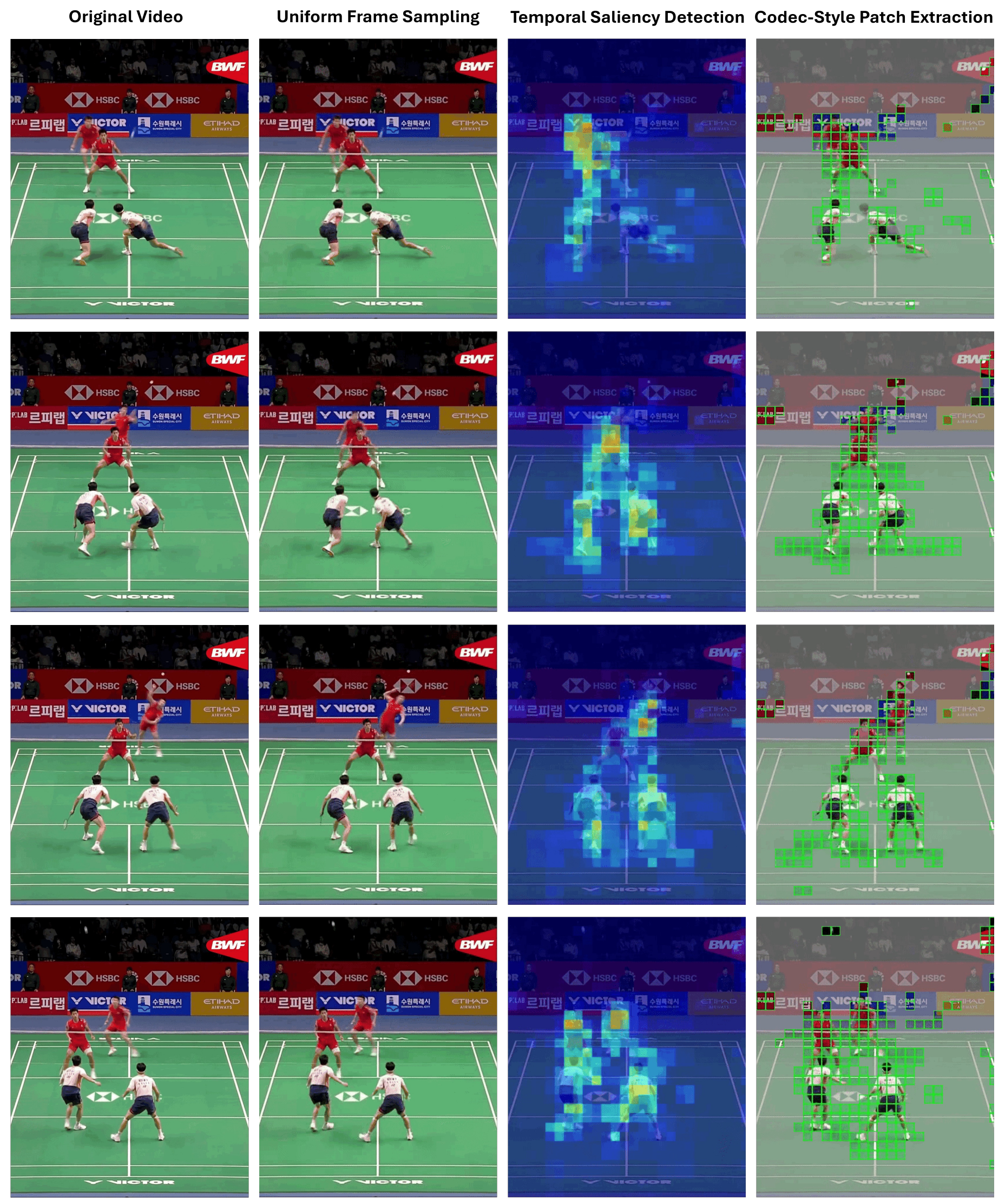}
    \caption{Comparison of video processing pipelines for spatiotemporal representation learning. (a) original dense video input with full temporal context, (b) uniform frame sampling that sparsely selects evenly spaced frames, (c) temporal saliency detection that identifies motion- and event-centric regions across all frames, and (d) Codec patch extraction that selectively retains temporally salient patches under a fixed token budget.}
\label{fig:case2}
\end{figure*}

\clearpage

{
    \small
    \bibliographystyle{ieeenat_fullname}

}

\end{document}